\RequirePackage{fix-cm}
\documentclass[twocolumn]{svjour3}          % twocolumn
\smartqed  % flush right qed marks, e.g. at end of proof
\usepackage{graphicx}
\usepackage{csvsimple}
\usepackage{amsmath}
\usepackage{cite}
\usepackage{url}
\usepackage[mathscr]{eucal}
\usepackage{amsfonts}
\usepackage{booktabs}
\newcommand{\subparagraph}{}

\usepackage[compact]{titlesec}
\titlespacing{\section}{1pt}{4ex}{2ex}
\titlespacing{\subsection}{1pt}{1ex}{1ex}
\titlespacing{\subsubsection}{1pt}{1ex}{1ex}

\begin{document}

\title{Biometrics Recognition Using Deep Learning: A Survey}


\author{Shervin Minaee, Amirali Abdolrashidi, Hang Su, Mohammed Bennamoun, David Zhang}


\institute{Shervin Minaee \at
              Snapchat, Machine Learning R\&D
              %\\ \email{fauthor@example.com}   
           \and
           Amirali Abdolrashidi \at
              University of California, Riverside
            \and
           Hang Su \at
              Facebook Research
            \and
           Mohammed Bennamoun \at
              The University of Western Australia
            \and
           David Zhang \at
              Chinese University of Hong Kong
}

\iffalse
\institute{F. Author \at
              first address \\
              Tel.: +123-45-678910\\
              Fax: +123-45-678910\\
              \email{fauthor@example.com}           %  \\
           \and
           S. Author \at
              second address
}
\fi

%%%\date{Received: date / Accepted: date}

\maketitle

\begin{abstract}
Deep learning-based models have been very successful in achieving state-of-the-art results in many of the computer vision, speech recognition, and natural language processing tasks in the last few years.
These models seem a natural fit for handling the ever-increasing scale of biometric recognition problems, from cellphone authentication to airport security systems. 
Deep learning-based models have increasingly been leveraged to improve the accuracy of different biometric recognition systems in recent years.
In this work, we provide a comprehensive survey of more than 120 promising works on biometric recognition (including face, fingerprint, iris, palmprint, ear, voice, signature, and gait recognition), which deploy deep learning models, and show their strengths and potentials in different applications. 
For each biometric, we first introduce the available datasets that are widely used in the literature and their characteristics. We will then talk about several promising deep learning works developed for that biometric, and show their performance on popular public benchmarks.
We will also discuss some of the main challenges while using these models for biometric recognition, and possible future directions to which research in this area is headed.
\end{abstract}

\section{Introduction}
\label{intro}
Biometric features\footnote{In this paper, we commonly refer to a biometric characteristic as \textit{biometric} for short.} hold a unique place when it comes to recognition, authentication, and security applications \cite{jain2000biometric}, \cite{zhang2013automated}. 
They cannot get lost, unlike token-based features such as keys and ID cards, and they cannot be forgotten, unlike knowledge-based features, such as passwords or answers to security questions \cite{zhang2018advanced}.
In addition, they are almost impossible to perfectly imitate or duplicate. 
Even though there have been recent attempts to generate and forge various biometric features \cite{galbally2008fake}, \cite{eskimez2018generating}, there have also been methods proposed to distinguish fake biometric features from authentic ones \cite{deepfake}, \cite{mo2018fake}, \cite{li2018exposing}. 
Changes over time for many biometric features are also extremely little. %(e.g. ear does not change significantly between age 7 to 70). 
For these reasons, they have been utilized in many applications, including cellphone authentication, airport security, and forensic science. Biometric features can be   \textit{physiological}, which are features possessed by any person, such as fingerprints \cite{jain2004introduction}, palmprints \cite{lu2003palmprint}, \cite{zhang1999two}, facial features \cite{face_Chellappa}, ears \cite{mu2004shape}, irises \cite{daugman2009iris}, \cite{bowyer2016handbook}, and retinas \cite{borgen_retina}, or \textit{behavioral}, which are apparent in a person's interaction with the environment, such as signatures \cite{elhoseny2018hybrid}, gaits \cite{gait_review}, and keystroke \cite{monrose2000keystroke}.
Voice/Speech contains both behavioral features, such as accent, and physiological features, such as voice pitch \cite{speech_rec}.

\begin{figure}[h]
\begin{center}
   \includegraphics[page=1,width=0.7\linewidth]{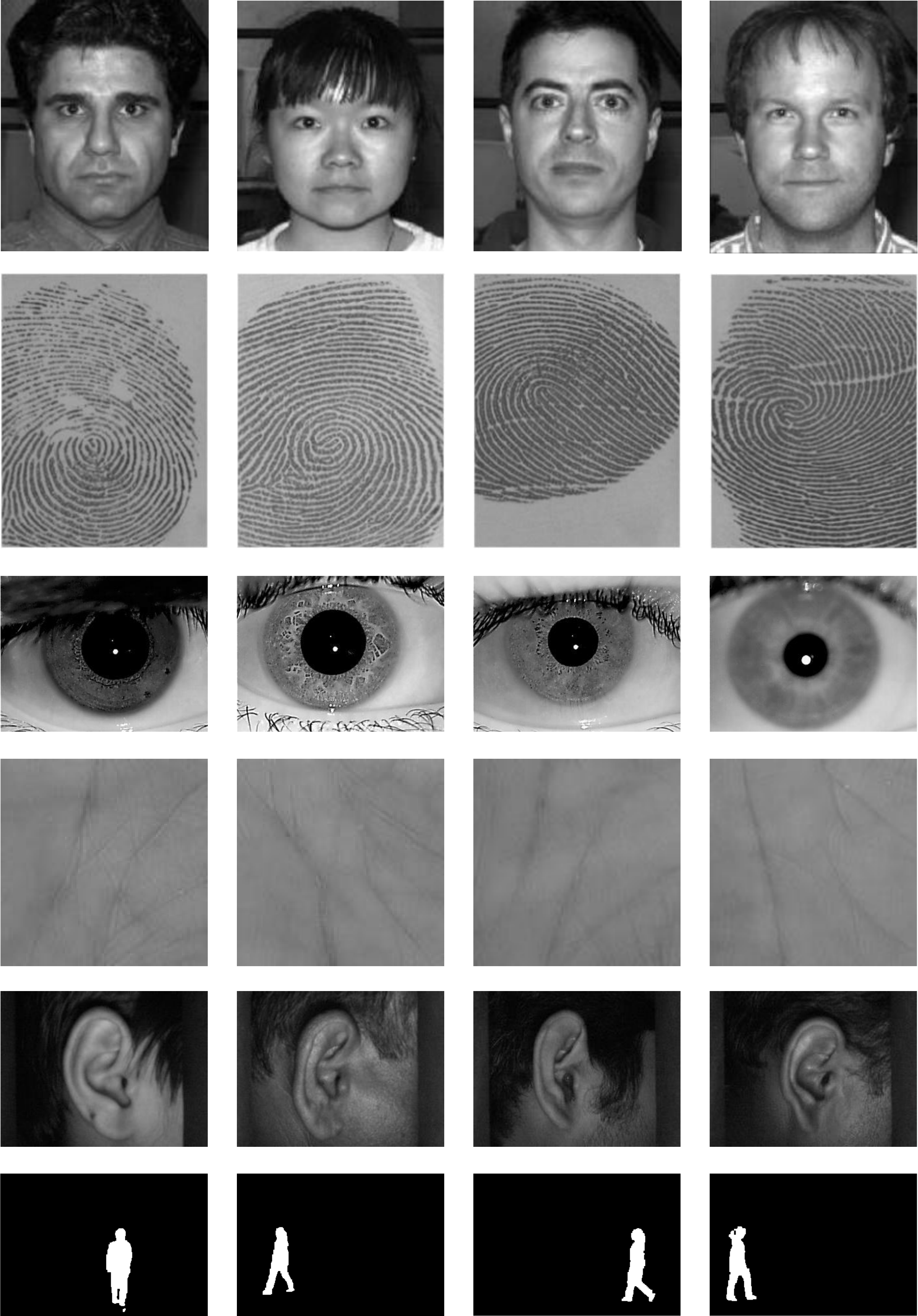}
\end{center}
   \caption{Sample images for various biometrics. The images in the first to sixth rows denote samples from face, fingerprint, iris, palmprint, ear, and gait respectively \cite{casia_gaitA, yaleb, polyU_finger, polyu_palm, kumar2010comparison, iit_ear}.}
\label{fig:biometrics_example}
\end{figure}

% Different biometrics
Face and fingerprint are arguably the most commonly used physiological biometric feature. Fingerprint is the oldest, dating back to 1893 when it was used to convict a murder suspect in Argentina \cite{hawthorne2017fingerprints}. 
Face has many discriminative features which can be used for recognition tasks \cite{jain2011handbook}. 
However, its susceptibility to change due to factors such as expression or aging, may present a challenge \cite{guo2016ei3d}, \cite{park2010age}.
Fingerprint consists of ridges and valleys, which form unique patterns. Minutiae are major local portions of the fingerprint which can be used to determine the uniqueness of the fingerprint, two of the most important ones of which being ridge endings and ridge bifurcations \cite{jain1997line}.
Palmprint is another alternative used for authentication purposes. In addition to minutiae features, palmprints also consist of geometry-based features, delta points, principal lines, and wrinkles \cite{multispectral_palm}. 
Iris and retina are the two most popular biometrics that are present in the eye, and can be used for recognition through the texture of the iris or the pattern of blood vessels in the retina. One interesting point worth noting is that even the two eyes in the same person have different patterns \cite{iris_survey_old}. Ears can also be used as a biometric through the shape of their lobes, and helix, and unlike most biometric features, do not need the person's direct interaction. The right and left ears are symmetrical in a person in most cases. However, their sizes are subject to change over time \cite{emervsivc2017ear}.
Among the behavioral features, signatures are arguably the most widely used today. The strokes in the signature can be examined for the pressure of the pen throughout the signature as well as the speed, which is a factor in the thickness of the stroke \cite{galbally2015line}. Gait refers to the manner of walking, which has been gaining more attention in the recent years. Due to the involvement of many joints and body parts in the process of walking, gait can also be used to uniquely identify a person from a distance \cite{wang2004fusion}. 
%Keystroke dynamics, i.e. how one types (typing rate, total time, etc.), can be used to distinguish authentic users from intruders posing as those users in a computer system \cite{morales2015keystroke}. However, this survey will not discuss keystroke dynamics.
% I have commented out the parts about keystroke dynamics. They are mentioned very few times (only up to here, in fact!), and we never explain or evaluate them in this paper.
Samples of various biometrics are shown in Figure \ref{fig:biometrics_example}.

\begin{figure}[h]
\begin{center}
   \includegraphics[page=2,width=0.99\linewidth]{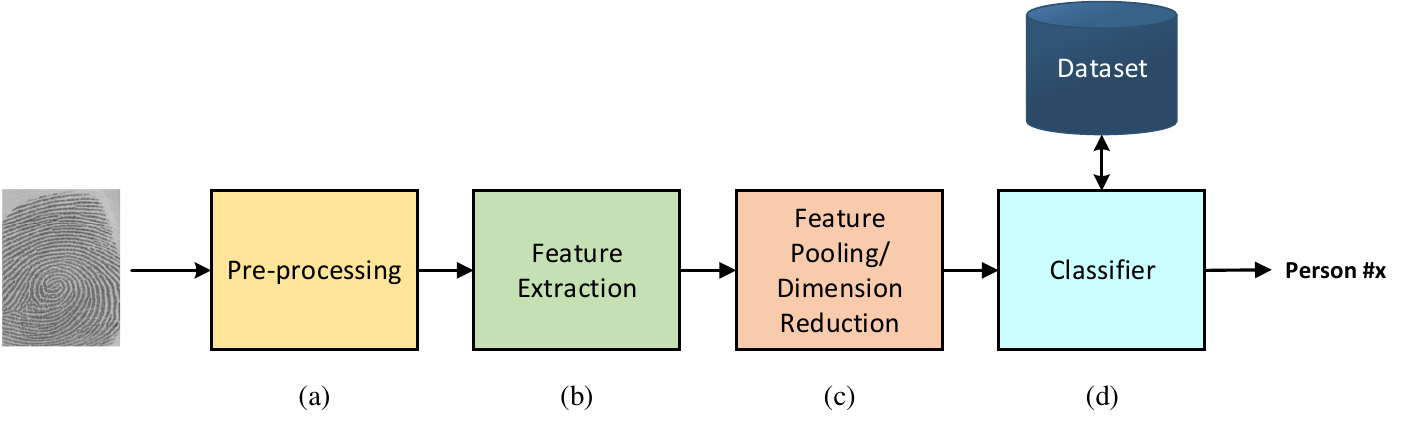}
\end{center}
   \caption{The block-diagram of most of classical biometric recognition algorithms.}
\label{fig:conventional_diagram}
\end{figure}

% Steps in traditional methods
Traditionally, the biometric recognition process involved several key steps. 
Figure \ref{fig:conventional_diagram} shows the block-diagram of traditional biometric recognition systems.
Firstly, the image data are acquired via (various) camera or optical sensors, and are then pre-processed so as to make the algorithm work on as much useful data as possible. 
%For example, the images are sometimes cropped and aligned to maximize the regions of interest (ROI).
Then, features are extracted from each image. 
Classical biometric recognition works were mostly based on hand-crafted features (designed by computer vision experts) to work with a certain type of data \cite{ahonen2004face}, \cite{2d_3d_face}, \cite{zhang2009advanced}.
Many of the hand-crafted features were based on the  distribution of edges (SIFT \cite{lowe2004distinctive}, HOG \cite{dalal2005histograms}), or where derived from transform domain, such as Gabor \cite{kong2003palmprint}, Fourier \cite{lai2001face}, and wavelet \cite{jin2004efficient}. 
Principal component analysis  is also used in many works to reduce the dimensionality of the features  \cite{abdi2010principal}, \cite{yang2004two}.
Once the features are extracted, they are fed into a classifier to perform recognition. 
%The classifier can be supervised, semi-supervised, or unsupervised.
%In supervised learning, the images are divided into training and testing images. The classifier uses the training images and their labels to tune its parameters and learn the tasks. 
%Then it will use the testing images to verify its accuracy \cite{kotsiantis2007supervised}.
%In unsupervised learning, the training labels are unknown. Therefore, it is up to the classifier to determine the clusters of data, so that the data within the cluster are closest to each other while being farthest from the data in the other clusters \cite{ng2006medical}. 
%In semi-supervised learning, both labeled and unlabeled data are used \cite{zhu2009introduction}.

% Challenges regarding traditional methods
Many challenges arise in a traditional biometric recognition task. 
For example, the hand-crafted features that are suitable for one biometric, will not necessarily perform well on others. 
Therefore, it would take a great number of experiments to find and choose the most efficient set of hand-crafted features for a certain biometric. 
Also many of the classical models were based on multi-class SVM trained in an one-vs-one fashion, which will not scale well when the number of classes is large.

\begin{figure*}[h]
\begin{center}
   \includegraphics[page=4,width=0.64\linewidth]{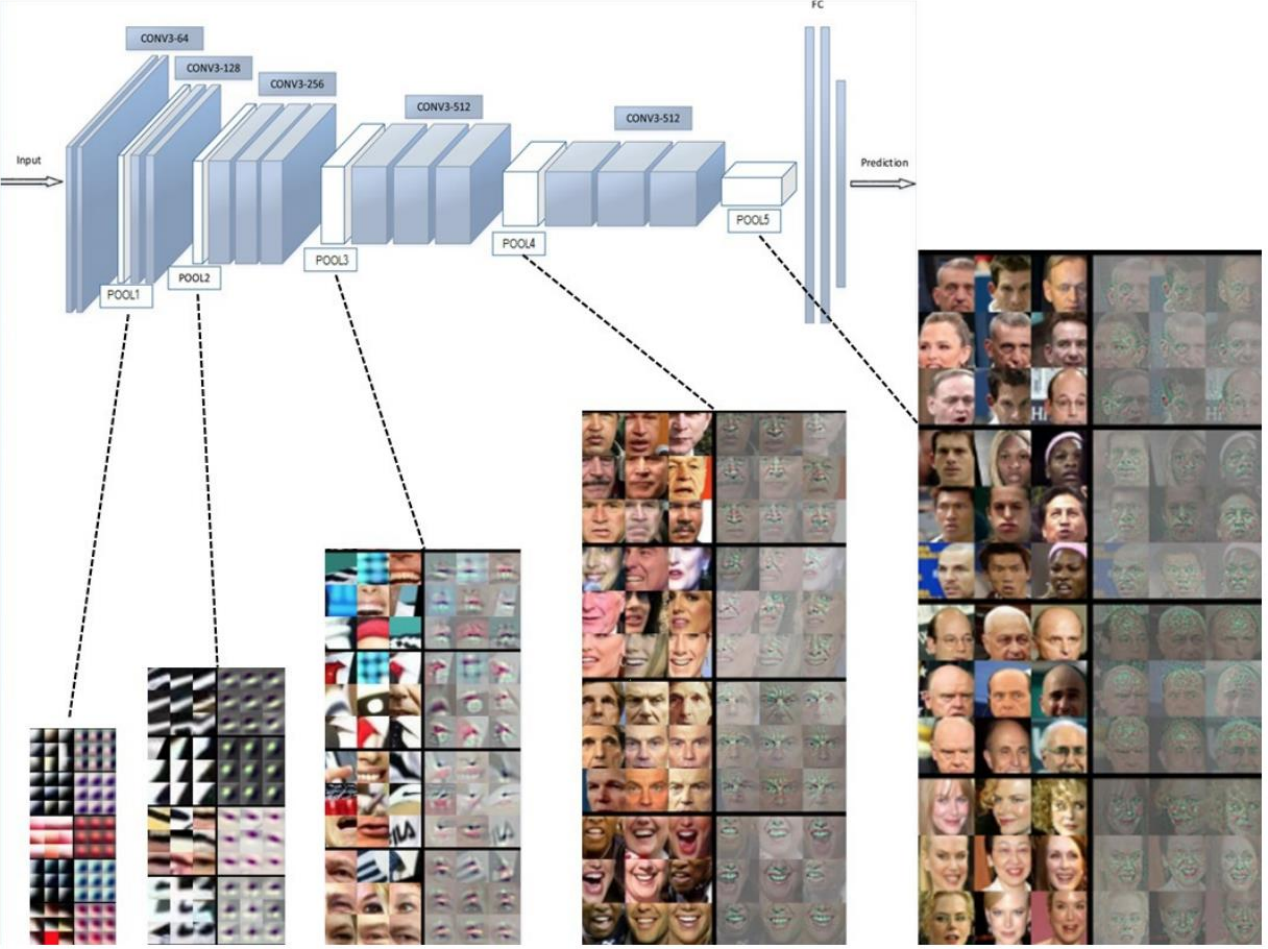}
\end{center}
   \caption{Illustration of the hierarchical concepts learned by a deep learning models trained for face recognition. Courtesy of \cite{deep_face_survey}.}
\label{face_hierarchy}
\end{figure*}

% Deep learning: A new beginning for vision and NLP
However, a paradigm shift started to occur in 2012, when a deep learning-based model, AlexNet \cite{alexnet}, won the ImageNet competition by a large margin.
Since then, deep learning models have been applied to a wide range of problems in computer vision and Natural Language Processing (NLP), and achieved promising results. 
Not surprisingly, biometric recognition methods were not an exception, and were taken over by deep learning models (with a few years delay).
Deep learning based models provide an end-to-end learning framework, which can jointly learn the feature representation while performing classification/regression.  This is achieved through a multi-layer neural networks, also known as \textit{Deep Neural Networks (DNNs)}, to learn multiple levels of representations that correspond to different levels of abstraction, which is better suited to uncover underlying patterns of the data (as shown in Figure \ref{face_hierarchy}). 
The idea of a multi-layer neural network dates back to the 1960s \cite{ivakhnenko1966cybernetic, schmidhuber2015deep}. However, their feasible implementation was a challenge in itself, as the training time would be too large (due to lack of powerful computers at that time). The progresses made in processor technology, and especially the development of General-Purpose GPUs (GPGPUs), as well as development of new techniques (such as Dropout) for training neural networks with a lower chance of over-fitting, enabled scientists to train very deep neural networks much faster \cite{lecun2015deep}.
The main idea of a neural network is to pass the (raw) data through a series of interconnected \textit{neurons} or \textit{nodes}, each of which emulates a linear or non-linear function based on its own \textit{weights} and \textit{biases}. 
These weights and biases would change during the training through back-propagation of the gradients from the output \cite{backprop}, usually resulted from the differences between the expected output and the actual current output, aimed to minimized a \textit{loss function} or \textit{cost function} (difference between the predicted and actual outputs according to some metric) \cite{bottou1991stochastic}. 
We will talk about different deep architectures in more details in Section 2.

Using deep models for biometric recognition, one can learn a hierarchy of concepts as we go deeper in the network.
Looking at face recognition for example, as shown in Figure \ref{face_hierarchy}, starting from the first few layers of the deep neural network, we can observe learned patterns similar to the Gabor feature (oriented edges with different scales). 
The next few layers can learn more complex texture features and part of the face.
The following layers are able to catch more complex pattern, such as high-bridged nose and big eyes.
Finally the last few layers can learn very abstract concepts and certain facial attribute (such as smile, roar, and even eye color faces). 

In this paper, we present a comprehensive review  of  the recent advances in biometric recognition using deep learning frameworks.
For each work, we provide an overview of the the key contributions, network architecture, and loss functions, developed to push state-of-the-art performance in biometric recognition.
We have gathered more than 150 papers, which appeared
between 2014 and 2019, in leading computer vision, biometric recognition, and machine learning conferences and journals.
For each biometric, we provide some of the most popular datasets used by the computer vision community, and the most promising state-of-the-art deep learning works utilized in the area of biometric recognition.
We then provide a quantitative analysis of well-known models for each biometric.
Finally, we  explore the challenges associated with deep learning-based methods in biometric recognition and research opportunities for the future.

The goal of this survey is to help new researchers in this field to navigate through the progress of deep learning-based biometric recognition models, particularly with the growing interest of multi-modal biometrics systems \cite{multimodal_challenge}.
Compared to the existing literature, the main contributions of this paper are as follow:
\begin{itemize}
    \item To the best of our knowledge, this is the only review paper which provides an overview of eight popular biometrics proposed before and in 2019, including face, fingerprint, iris, palmprint, ear, voice, signature, and gait.
    \item We cover the contemporary literature with respect to this area. We present a comprehensive review of more than 150 methods, which have appeared since 2014.
    \item We provide a comprehensive review and an insightful analysis of different aspects of biometric recognition using deep learning, including the training data, the choice of network architectures, training strategies, and their key contributions.
    \item We provide a comparative summary of the properties and performance of the reviewed methods
    for biometric recognition.
    %\item We provide state-of-the-art algorithms for these 8 biometrics, which can can help researchers to easily navigate through those works and get their main idea. 
    \item We provide seven challenges and potential future direction for deep learning-based biometric recognition models.
\end{itemize}

The structure of the rest of this paper is as follows. 
In Section 2, we provide an overview of popular deep neural networks architectures, which serve as the backbone of many biometric recognition algorithms, including convolutional neural networks, recurrent neural networks, auto-encoders, and generative adversarial networks. 
Then in Section 3, we provide an introduction to each of the eight biometrics (Face, Fingerprint, Iris, Palmprint, Ear, Voice, Signature, and Gait), some of the popular datasets for each of them, as well as the promising deep learning based works developed for them.
The quantitative results and experimental performance of these models for all biometrics are provided in Section 4. Finally in Section 5, we explore the challenges and future directions for deep learning-based biometric recognition.

\section{Deep Neural Network Overview}
\label{sec:DNNs}
In this section, we provide an overview of some of the most promising deep learning architectures used by the computer vision community, including convolutional neural networks (CNN) \cite{CNN}, recurrent neural networks (RNN) and one of their specific version called long short term memory (LSTM) \cite{lstm}, auto-encoders, and generative adversarial networks (GANs) \cite{gan}.
It is noteworthy that with the popularity of deep learning in recent years, there are several other deep neural architectures proposed (such as Transformers, Capsule Network, GRU, and spatial transformer networks), which we will not cover in this work.

\begin{figure}[h]
\begin{center}
   \includegraphics[page=3,width=0.98\linewidth]{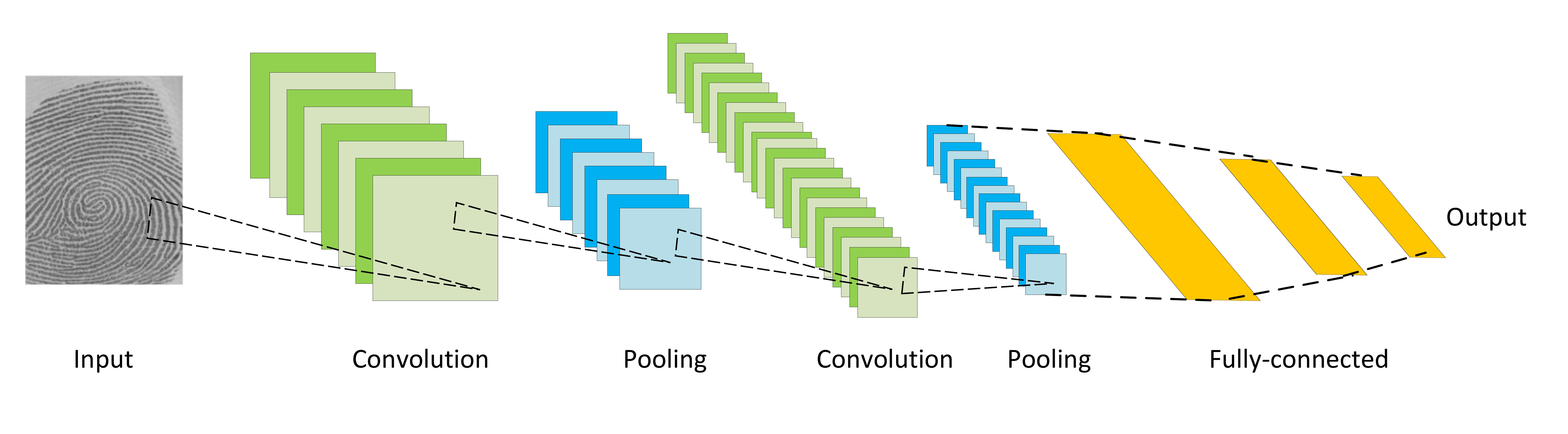}
\end{center}
   \caption{Architecture of a Convolutional Neural Network (CNN), showing the main two operations of convolution and pooling. Courtesy of Yann LeCun.}
\label{fig:CNN_arch}
\end{figure}

\textbf{\subsection{Convolutional Neural Networks (CNN)}}
Convolutional Neural Networks (CNN) (inspired
by the mammalian visual cortex) are one of the most successful and widely used architectures in deep learning community (specially for computer vision tasks).
CNN was initially proposed by Fukushima in a seminal paper, called "Neocognitron" \cite{neocog}, based on the model of human visual system proposed by Nobel laureates Hubel and Wiesel. Later on Yann Lecun and colleagues developed an optimization framework (based on back-propagation) to efficiently learn the model weights for a CNN architecture \cite{CNN}.
The block-diagram of one of the first CNN models developed by Lecun et al. is shown in Figure \ref{fig:CNN_arch}.

CNNs mainly consist of three type of layers: convolutional layers, where a sliding kernel is applied to the image (as in image convolution operation) in order to extract features; nonlinear layers (usually applied in an element-wise fashion), which apply an activation function on the features in order to enable the modeling of non-linear functions by the network; and pooling layers, which takes a small neighborhood of the feature map and replaces it with some statistical information (mean, max, etc.) of the neighborhood.
Nodes in the CNN layers are locally connected;
that is, each unit in a layer receives input from a small neighborhood of the previous layer (known as the receptive field). 
The main advantage of CNN is the weight sharing mechanism through the use of the sliding kernel, which goes through the images, and aggregates the local information to extract the features.
Since the kernel weights are shared across the entire image, CNNs have a significantly smaller number of parameters than a similar fully connected neural network.
Also by stacking multiple convolution layers, the higher-level layers learn features from increasingly wider receptive fields. 

CNNs have been applied to various computer vision tasks such as: semantic segmentation \cite{seman_seg}, medical image segmentation \cite{med_seg}, object detection \cite{faster_rcnn}, super-resolution \cite{sisr}, image enhancement \cite{enhance}, caption generation for image and videos \cite{caption}, and many more. 
Some of the most well-known CNN architectures include AlexNet \cite{alexnet}, ZFNet \cite{zfnet}, VGGNet \cite{vggnet}, ResNet \cite{resnet}, GoogLenet \cite{googlenet}, MobileNet \cite{mobilenet}, and DenseNet \cite{densenet}. 
%These models can be used as the backbone network for many of the transfer learning-based models.

\begin{figure}[h]
\begin{center}
   \includegraphics[page=5,width=0.9\linewidth]{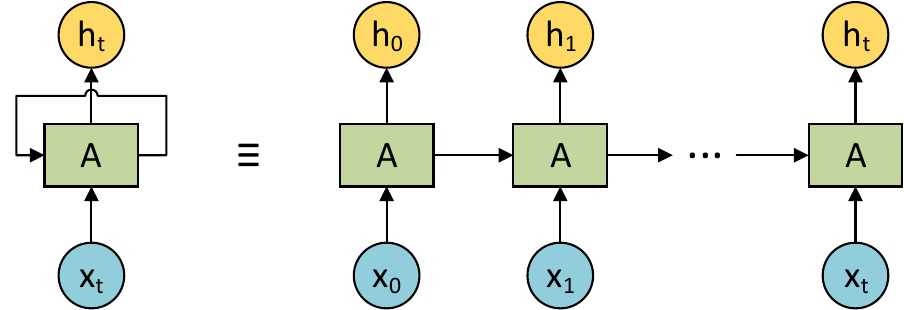}
\end{center}
   \caption{Architecture of a Recurrent Neural Network (RNN).}
\label{fig:RNN_arch}
\end{figure}

\textbf{\subsection{Recurrent Neural Networks and LSTM}}
Recurrent Neural Networks (RNNs) \cite{RNN} are widely used for processing sequential data like speech, text, video, and time-series (such as stock prices), where data at any given time/position depends on the previously encountered data.
A high-level architecture of a simple RNN is shown in Figure \ref{fig:RNN_arch}. As we can see at each time-stamp, the model gets the input from the current time $X_i$ and the hidden state from the previous step $h_{i-1}$ and outputs the hidden state (and possibly an output value). The hidden state from the very last time-stamp (or a weighted average of all hidden states) can then be used to perform a task.

RNNs usually suffer when dealing with long sequences, as they cannot capture the long-term dependencies of many real application (although in theory there is nothing limiting them from doing so). However, there is a variation of RNNs, called LSTM, which is designed to better capture long-term dependencies. 

\begin{figure}[h]
\begin{center}
   \includegraphics[page=14,width=0.64\linewidth]{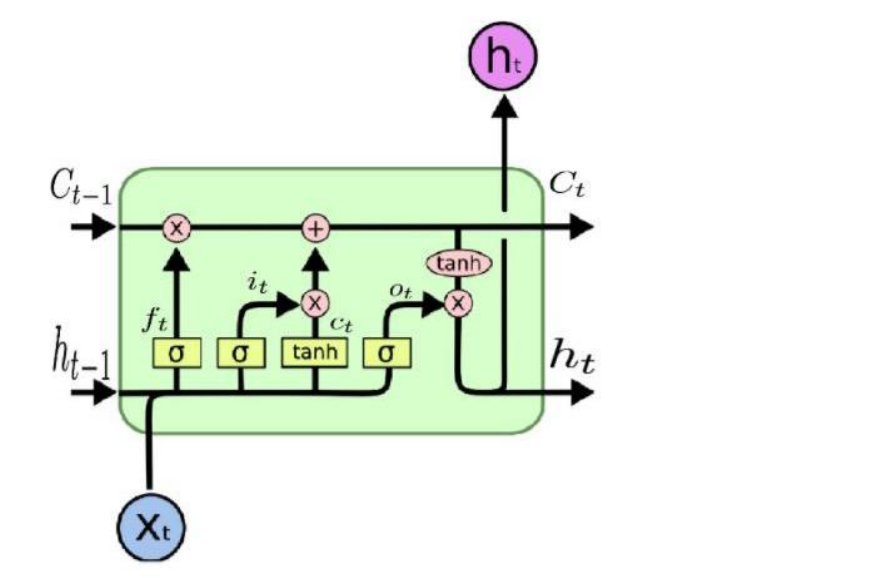}
\end{center}
   \caption{The architecture of a standard LSTM module, courtesy of Andrej Karpathy \cite{lstm_cell}.}
\label{fig:lstm_model}
\end{figure}

\textbf{Long Short Term Memory (LSTM):}
LSTM is a popular recurrent neural network architecture for modeling sequential data, which is designed to have a better ability to capture long term dependencies than the vanilla RNN model \cite{lstm}.
%Similar to  other recurrent neural networks, At each time-step LSTM gets the input from the current time-step and the hidden state from the previous time-step, and produces an output which is fed to the next time step.  
As mentioned above, the vanilla RNN often suffers from the gradient vanishing or exploding problems, and LSTM network tries to overcome this issue by introducing some internal gates.
In the LSTM architecture, there are three gates (input gate, output gate, forget gate) and a memory cell.
The cell remembers values over arbitrary time intervals and the other three gates regulate the flow of information into and out of the cell.
Figure \ref{fig:lstm_model} illustrates the inner architecture of a single LSTM module.

The relationship between input, hidden states, and different gates is shown in Equation \ref{eq_lstm}:
\begin{equation}
\label{eq_lstm}
\begin{aligned}
%\centering
f_t &= \sigma (\textbf{W}^{(f)} x_t+\textbf{U}^{(f)} h_{t-1}+ b^{(f)}  )\\
i_t &= \sigma (\textbf{W}^{(i)} x_t+\textbf{U}^{(i)} h_{t-1}+ b^{(i)}  ) \\
o_t &= \sigma (\textbf{W}^{(o)} x_t+\textbf{U}^{(o)} h_{t-1}+ b^{(o)}  ) \\
c_t &= f_t \odot c_{t-1} \\
& + i_t \odot \text{tanh} (\textbf{W}^{(c)} x_t+\textbf{U}^{(c)} h_{t-1}+ b^{(c)}  ) \\
h_t &= o_t \odot \text{tanh}(c_t)
\end{aligned}
\end{equation}
where $x_t \in R^d$ is the input at time-step t, and $d$ denotes the feature dimension for each word, $\sigma$ denotes the element-wise sigmoid function (to squash/map the values within $[0,1]$),  $\odot$ denotes the element-wise product.
$c_t$ denotes the memory cell designed to lower the risk of vanishing/exploding gradient, and therefore enabling the learning of dependencies over larger periods of time, which is infeasible with traditional recurrent networks.
The forget gate, $f_t$ is to reset the memory cell. 
$i_t$ and $o_t$ denote the input and output gates, and essentially control the input and output of the memory cell.
%For many applications, we are interested in the temporal information flow in both directions, and there is variant of LSTM, called Bidirectional-LSTM (Bi-LSTM), which can address this. Bidirectional LSTMs train two hidden layers on the input sequence. The first one on the input sequence as-is, and the second one on the reversed copy of the input sequence. This can provide additional context to the network, by looking at both past and future information, and results in faster and better learning. LSTMs have been widely used in NLP, speech, and vision applications.

\begin{figure}[h]
\begin{center}
   \includegraphics[page=4,width=0.9\linewidth]{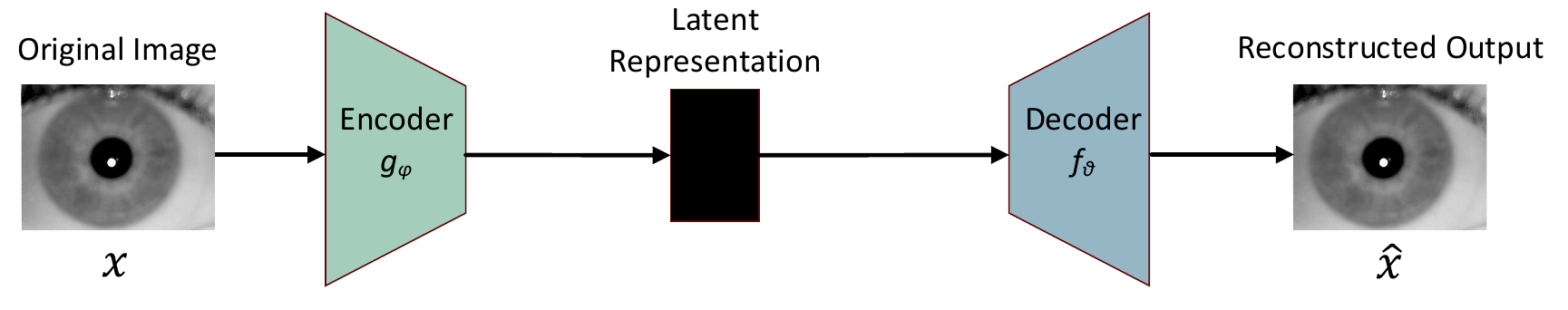}
\end{center}
   \caption{Architecture of a standard Auto-Encoder Model.}
\label{fig:autoencoder}
\end{figure}

\textbf{\subsection{Auto-Encoders}}
Auto-encoders are a family of neural network models used to learn efficient data encoding in an unsupervised manner. They achieve this by compressing the input data into a latent-space representation, and then reconstructing the output (which is usually the same as the input) from this representation. 
Auto-encoders are composed of two parts:
\\ \textbf{Encoder:} 
This is the part of the network that compresses the input into a latent-space representation. It can be represented by an encoding function $z=f(x)$.
\\ \textbf{Decoder:} 
This part aims to reconstruct the input from the latent space representation. It can be represented by a decoding function $y=g(z)$.
\\The architecture of a simple auto-encoder model is demonstrated in Figure \ref{fig:autoencoder}.
Auto-encoders are usually trained by minimizing the reconstruction error, $L(x,\hat{x})$ (unsupervised, i.e. no need for labeled data), which measures the differences between our original input $x$, and the consequent reconstruction $\hat{x}$.
Mean square error, and mean absolute deviation are popular choices for the reconstruction loss in many applications. 
%One can also add some sparsity regularization on the latent vector to promote data compression.

There are several variations of auto-encoders where were proposed in the past.
One of the popular ones is the stacked denoising auto-encoder (SDAE) which stacks several auto-encoders and uses them for image denoising \cite{stac_ae}.
Another popular variation of autoencoders is "variational auto-encoder (VAE)" which imposes a prior distribution on the latent representation \cite{vr_ae}. Variational auto-encoders are able to generate realistic samples from a data distribution.
Another variation of auto-encoders is the adversarial auto-encoders, which introduces an adversarial loss on the latent representation to encourage them to be close to a prior distribution.
%Adversarial auto-encoders have also been used for unsupervised feature learning from medical images too \cite{baf}.

\begin{figure}[h]
\begin{center}
   \includegraphics[page=1,width=0.99\linewidth]{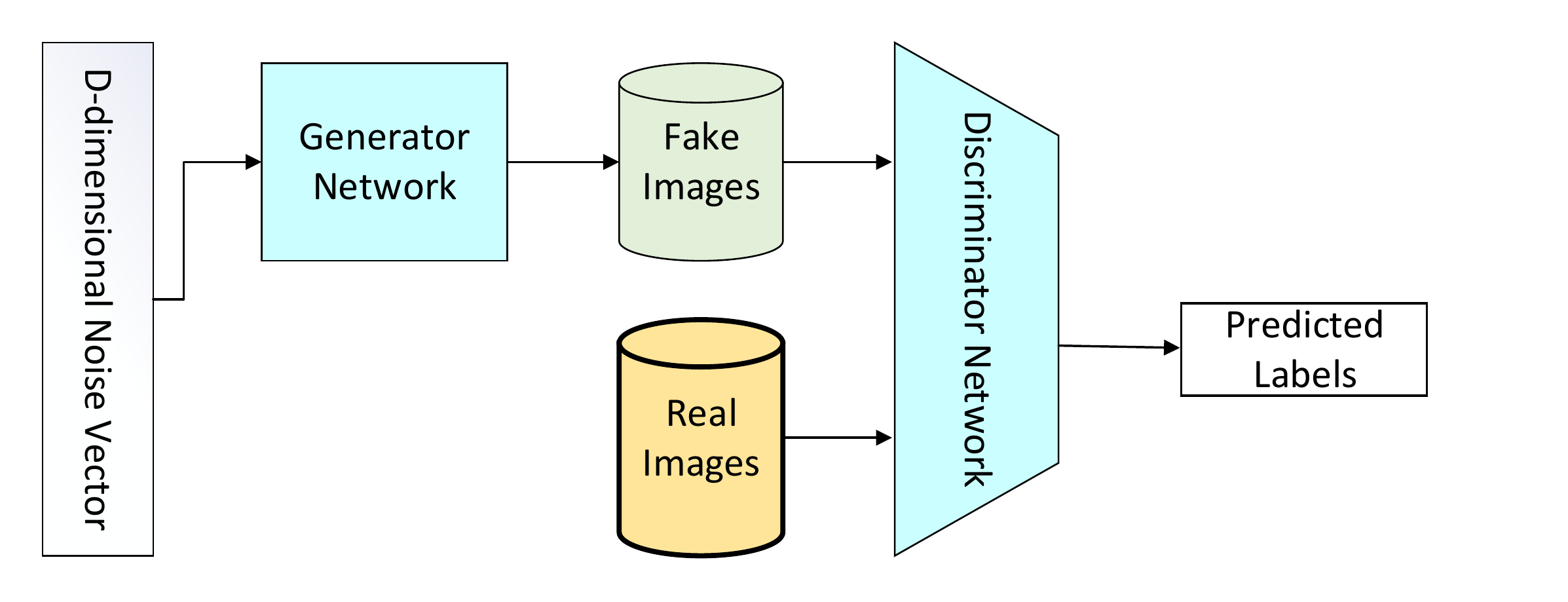}
\end{center}
   \caption{The architecture of generative adversarial network.}
\label{fig:gen_arch}
\end{figure}

\textbf{\subsection{Generative Adversarial Networks (GAN)}}
Generative Adversarial Networks (GANs) are a newer family of deep learning models, which consists of two networks, one generator, and one discriminator \cite{gan}. 
On a high level, the generator's job is to generate samples from a distribution which are close enough to real samples with the objective to fool the discriminator, while the discriminator's job is to distinguish the generated samples (fakes) from the authentic ones. 
The general architecture of a vanilla GAN model is demonstrated in Figure \ref{fig:gen_arch}.

The generator network in vanilla GAN learns a mapping from noise $z$ (with a prior distribution, such as Gaussian) to a target distribution $y$, $G= z \rightarrow y$,  which look similar to the real samples, while the discriminator network, $D$, tries to distinguish the samples generated by the generator models from the real ones.
The loss function of GAN can be written as Equation \ref{eq_gan}:
\begin{small}
\begin{equation}
\begin{aligned}
\mathcal{L}_{GAN} &=  \mathbb{E}_{x \sim p_{data}(x)}[\text{log} D(x)]\\ &+ \mathbb{E}_{z \sim p_z(z)}[\text{log}(1-D(G(z)))]
\end{aligned}
\label{eq_gan}
\end{equation}
\end{small}
in which we can think of GAN, as a minimax game between $D$ and $G$, where $D$ is trying to minimize its classification error in detecting fake samples from the real ones (maximize the above loss function), and $G$ is trying to maximize the discriminator network's error (minimize the above loss function).
After training this model, the trained generator model would be: 
\begin{equation}
\begin{aligned}
G^*= \text{arg} \ \text{min}_G \text{max}_D \ \mathcal{L}_{GAN}
\end{aligned}
\label{mini_max}
\end{equation}
In practice, the loss function in Equation \ref{mini_max}  may not provide enough gradient for $G$ to get trained well, specially at the beginning where $D$ can easily detect fake samples from the real ones. 
One solution is to maximize $\mathbb{E}_{z \sim p_z(z)}[\text{log}(D(G(z)))]$. %(instead of minimizing $\mathbb{E}_{z \sim p_z(z)}[\text{log}(1-D(G(z)))]$).

Since the invention of GAN, there have been several works trying to improve/modify GAN in different aspects.
%To name a few promising works, in \cite{dc-gan}, Radford et al. proposed a convolutional GAN model for generating images, which works better than fully connected networks when used for image generation. In \cite{con-gan}, Mirza proposed a conditional GAN model, which can generate images conditioned on class labels. This is a nice modification of the vanilla GAN model, which enables one to generate samples with a specified label. The work in \cite{was-gan} proposed a new loss function based on Wasserstein distance (also known as earth mover's distance), to better estimate the distance for cases where the distribution of the generated samples and the real ones are non-overlapping (for which KL divergence is not a good representative of the distance). 
%In \cite{cycle-gan}, Zhu proposed an image to image translation model based on a cycle-consistent GAN model, which learns to map a given image distribution into a target domain (e.g. day to night images, summer to winter images, horse to zebra, etc).
For a detailed list of works relevant to GAN, please refer to \cite{GanZoo}.

\textbf{\subsection{Transfer Learning Approach}}
Now that we talked about some of the popular deep learning architectures, let us briefly talk about how these models are applied to new applications.
Of course, these models can always be trained from scratch on new applications, assuming they are provided with sufficient labeled data. 
But depending on the depth of the model (i.e. how large if the number of parameters), it may not be very straightforward to make the model converge to a good local minimum. Also, for many applications, there may not be enough labeled data available to train a deep model from scratch.
For these situations,  transfer learning approach can be used to better handle labeled data limitations and the local-minimum problem. 

In transfer learning, a model trained on one task is re-purposed on another related task, usually by some adaptation toward the new task.
For example, one can imagine using an image classification model trained on ImageNet, to be used for a different task such as texture classification, or iris recognition.
There are two main ways in which the pre-trained model is used for a different task. 
In one approach, the pre-trained model, e.g. a language model, is treated as a feature extractor, and a classifier is trained on top of it to perform classification (e.g. sentiment classification). Here the internal weights of the pre-trained model are not adapted to the new task. 
In the other approach, the whole network, or a subset of it, is fine-tuned on the new task. Therefore the pre-trained model weights are treated as the initial values for the new task, and are updated during the training stage.

Many of the deep learning-based models for biometric recognition are based on transfer learning (except for voice because of the difference in the nature of the data, and face because of the availability of large-scale datasets), which we are going to explain in the following section.
\iffalse
Figure \ref{fig:vgg_tran} shows the block-diagram of a sample transfer learning approach, in which a pre-trained VGG Network (on ImageNet) is used for face recognition.
\begin{figure}[h]
\begin{center}
   \includegraphics[page=24,width=0.95\linewidth]{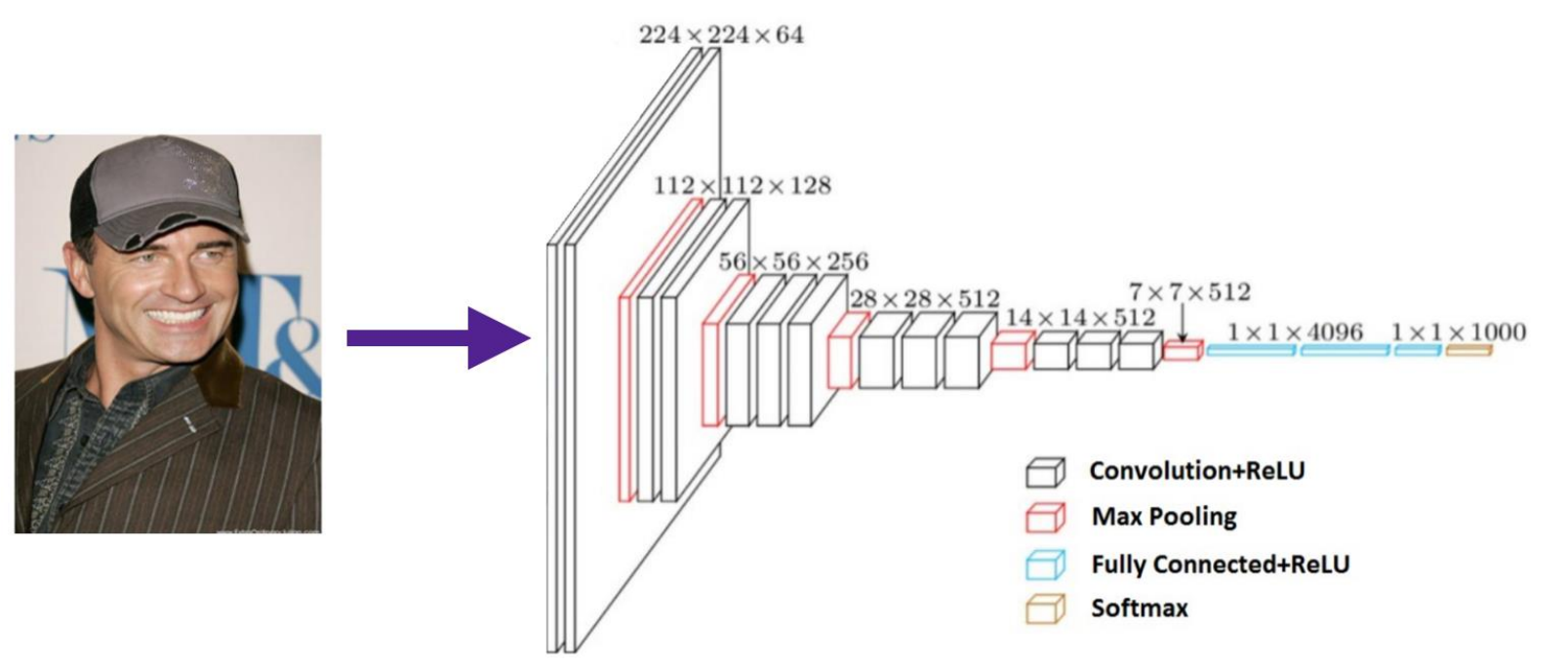}
\end{center}
   \caption{Face recognition using a pre-trained CNN.}%, courtesy of, Amir please draw a similar architecture to this one.}
\label{fig:vgg_tran}
\end{figure}
\fi

\section{Deep Learning Based Works on Biometric Recognition}
\label{sec:Deep}
In this section we provide an overview of some of the most promising deep learning works for various biometric recognition works. Within each subsection, we also provide a summary of some of the most popular datasets for each biometric.

\textbf{\subsection{Face Recognition}}
Face is perhaps one of the most popular biometrics (and the most researched one during the last few years). 
It has a wide range of applications, from security cameras in airports and government offices, to daily usage for cellphone authentication (such as in FaceID in iPhones). 
Various hand-crafted features were used for recognition in the past, such as the LBP, Gabor Wavelet, SIFT, HoG, and also sparsity-based representations \cite{deng2013defense}, \cite{cao2013similarity},  \cite{face_sparse},  \cite{yang2012regularized},  \cite{yi2013towards}.
Both 2D and 3D versions of faces are used for recognition \cite{mian2007efficient}, but most people have focused on 2D face recognition so far. 
One of the main challenges for facial recognition is the face's susceptibility to change over time due to aging or external factors, such as scars, or medical conditions \cite{park2010age}.
We will introduce some of the most widely used face recognition datasets in the next section, and then talk about the promising deep learning-based face recognition models.

\textbf{\subsubsection{Face Datasets}}
Due to the wide application of face recognition in the industry, a large number of datasets are proposed for that purpose. We will introduce some of the most popular ones here.

\begin{figure}[h]
\begin{center}
   \includegraphics[page=25,width=0.8\linewidth]{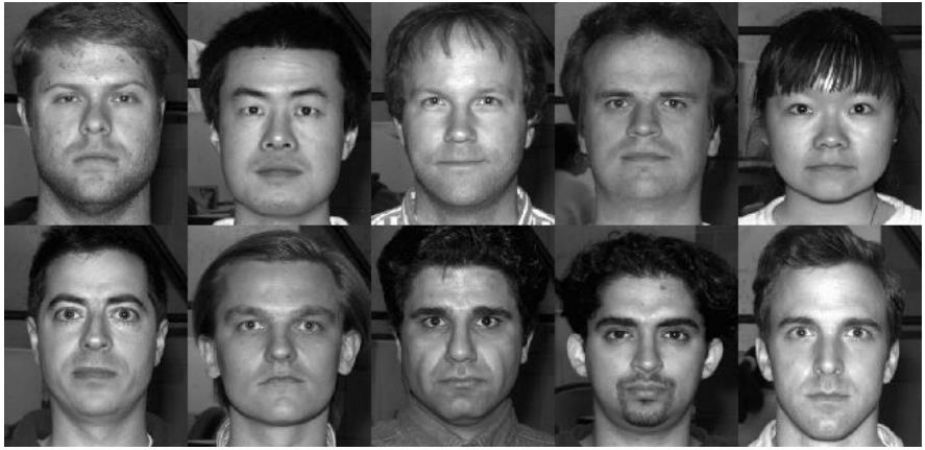}
\end{center}
   \caption{Ten example images from Yale Face B Dataset.}%, courtesy of, Amir please draw a similar architecture to this one.}
\label{fig:yale_face}
\end{figure}

\textbf{Yale and Yale Face Database B:}
Yale face dataset is perhaps one of the earliest face recognition datasets \cite{yale}. It Contains 165 grayscale images of 15 individuals. There are 11 images per subject, one per different facial expression or configuration (center-light, w/glasses, happy, left-light, w/no glasses, normal, right-light, sad, sleepy, surprised, and wink).
\\It is extended version, Yale Face Database B \cite{yaleb}, contains 5760 single light source images of 10 subjects each seen under 576 viewing conditions (9 poses x 64 illumination conditions). For every subject in a particular pose, an image with ambient (background) illumination was also captured.
Ten example images from Yale face B dataset are shown in Figure \ref{fig:yale_face}.

\textbf{CMU Multi-PIE:}
The CMU Multi-PIE face database contains more than 750,000 images of 337 people \cite{cmu}, \cite{gross2010multi}. 
Subjects were imaged under 15 view points and 19 illumination conditions while displaying a range of facial expressions. 

\textbf{Labeled Face in The Wild (LFW):}
Labeled Faces in the Wild is a database of face images designed for studying unconstrained face recognition. The database contains more than 13,000 images of faces collected from the web. Each face has been labeled with the name of the person pictured \cite{lfw}. 
1680 of the people pictured have two or more distinct photos in the database. 
The only constraint on these faces is that they were detected by the Viola-Jones face detector. For more details on this dataset, we refer the readers to the database web-page.

\textbf{PolyU NIR Face Database:}
The Biometric Research Centre at The Hong Kong Polytechnic University developed a NIR face capture device and used it to construct a large-scale NIR face database \cite{polyu_face}.
By using the self-designed data acquisition device, they collected NIR face images from 335 subjects. In each recording, 100 images from each subject is captured, and in total about 34,000 images were collected in the PolyU-NIRFD database.

\textbf{YouTube Faces:}
This data set contains 3,425 videos of 1,595 different people. All videos were downloaded from YouTube. An average of 2.15 videos are available for each subject. 
The goal of this dataset was to produce a large scale collection of videos along with labels indicating the identities of a person appearing in each video \cite{youtube}. 
In addition, they published benchmark tests, intended to measure the performance of video pair-matching techniques on these videos.

\textbf{VGGFace2:}
VGGFace2 is a large-scale face recognition dataset \cite{vggface2}. 
Images are downloaded from Google Image Search and have a large variations in pose, age, illumination, ethnicity and profession.
It contains 3.31 million images of 9131 subjects (identities), with an average of 362.6 images for each subject. Face distribution for different identities is varied, from 87 to 843.

\textbf{CASIA-WebFace:}
CASIA WebFace Facial dataset of 453,453 images over 10,575 identities after face detection \cite{casiawebface}. 
This is one of the largest publicly available face datasets.

\textbf{MS-Celeb:}
Microsoft Celeb is a dataset of 10 million face images harvested from the Internet for the purpose of developing face recognition technologies, from nearly 100,000 individuals \cite{msceleb1m}. 
%Microsoft's goal in building this dataset was to distribute an initial training dataset of 100,000 individuals' biometric data to accelerate research into recognizing a larger target list of one million people "using all the possibly collected face images of this individual on the web as training data.

\textbf{CelebA:}
CelebFaces Attributes Dataset (CelebA) is a large-scale face attributes dataset with more than 200K celebrity images \cite{celeba}.
%The images in this dataset cover large pose variations and background clutter. 
CelebA has a large diversity, large quantities, and rich annotations, including more than 10,000  identities, more than 202,599 face images, 5 landmark locations, 40 binary attributes annotations per image.
The dataset can be employed as the training and test sets for the following computer vision tasks: face attribute recognition, face detection, landmark (or facial part) localization, and face editing \& synthesis.

\textbf{IJB-C:}
The IJB-C dataset \cite{ijbc} contains about 3,500 identities with a total of 31,334 still facial images and 117,542 unconstrained video frames. 
The entire IJB-C testing protocols are designed to test detection, identification, verification and clustering of faces. In the 1:1 verification protocol, there are 19,557 positive matches and 15,638,932 negative matches.

\textbf{MegaFace:}
MegaFace Challenge \cite{megaface} is a publicly available  benchmark, which is widely used to test the performance of facial recognition algorithms (for both identification and verification). 
The gallery set of MegaFace contains over 1 million images from 690K identities collected from Flickr \cite{flicker}.
The probe sets are two existing databases: FaceScrub and FGNet. The FaceScrub dataset contains 106,863 face images of 530 celebrities. The FGNet dataset is mainly used for testing age invariant face recognition, with 1002 face images from 82 persons.

\textbf{Other Datasets:} 
It is worth mentioning that there are several other datasets which we skipped the details due to being private or less popularity, such as  DeepFace (Facebook private dataset of 4.4M photos of 4k subjects), NTechLab (a private dataset of 18.4M photos of 200k subjects), FaceNet (Google private dataset of more than 500M photos of more than 10M subjects), WebFaces (a dataset of 80M photos crawled from web) \cite{megaface}, and  Disguised Faces in the Wild (DFW) \cite{DFW} which contains over 11,000 images of 1,000 identities with variations across different types of disguise accessories.

\textbf{\subsubsection{Deep Learning Works on Face Recognition}}
There are countless number of works using deep learning for face recognition. 
In this survey, we provide an overview of some of the most promising works developed for face verification and/or identification. 

In 2014, Taigman and colleagues proposed one of the earliest deep learning work for face recognition in a paper called DeepFace \cite{DeepFace}, and achieved the state-of-the-art accuracy on the LFW benchmark \cite{lfw}, approaching human performance on the unconstrained condition for the first time ever (DeepFace: 97.35\% vs. Human: 97.53\%). 
DeepFace was trained on 4 million facial images.
This work was a milestone on face recognition, and after that several researchers started using deep learning for face recognition.

In another promising work in the same year, Sun et al.  proposed DeepID (Deep hidden IDentity features) \cite{DeepID}, for face verification.
DeepID features were taken from the last hidden layer  of a deep convolutional network, which is trained to   recognize about 10,000 face identities in the training set. 
%Although these features are learned through identification, the authors show them to be effective for face verification and new unseen faces.

In a follow up work, Sun et al. extended DeepID for joint face identification and verification called DeepID2 \cite{DeepID2}.
By training the model for joint identification and verification, they showed that the face identification task increases the inter-personal variations by drawing DeepID2 features extracted from different identities apart, while the face verification task reduces the intra-personal variations by pulling DeepID2 features extracted from the same identity together. %both of which are essential to face recognition.
For identification, cross-entropy is used as the loss function (as defined in the Equation \ref{eqn:eq_iden}), while for verification they proposed to use the loss function of Equation \ref{eqn:eq_verif} to reduce the intra-class distances on the features and increase the inter-class distances.
\begin{equation}
\mathbf{L}_{Ident} (f,t,\theta_{id})= - \sum_i p_i \log \hat{p}_i
\label{eqn:eq_iden}
\end{equation}

\begin{equation}
\begin{split}
\mathbf{L}_{Verif}& (f_i, f_j, y_{ij}, \theta_{vr})=
\\
&
\begin{cases}
    \frac{1}{2} \|f_i-f_j\|_2^2 ,& \text{if } y_{ij}=1\\
    \frac{1}{2} max(1- \frac{1}{2} \|f_i-f_j\|_2^2, 0),
    & \text{otherwise}
\end{cases}
\end{split}
\label{eqn:eq_verif}
\end{equation}
As an extension of DeepID2, in DeepID3 \cite{DeepID3} Sun et al proposed a new model which has higher dimensional hidden representation, and  deploys VGGNet and GoogleNet as the main architectures.

In 2015, FaceNet \cite{FaceNet} trained a GoogLeNet model on a large private dataset.  
This work tried to learn a mapping from face images to a compact Euclidean space where distances  directly corresponds to a measure of face similarity.
It adopted a triplet loss function based on triplets of roughly aligned matching/non-matching face patches generated by a novel online triplet mining method and achieved good performance on LFW dataset (99.63\%).
Given features for a given sample $f(x_i^a)$, a positive sample $f(x_i^p)$ (matching $x_i^a$), and a negative sample $f(x_i^n)$, the triplet loss for a given margin $\alpha$ is defined as Equation \ref{eq_triplet}:
\begin{equation}
\mathbf{L}_{triplet}= \sum \bigg[ \|f(x_i^a)-f(x_i^p)\|_2^2 - \|f(x_i^a)-f(x_i^n)\|_2^2+ \alpha \bigg]_+
\label{eq_triplet}
\end{equation}
In the same year, Parkhi et al. proposed a model called VGGface \cite{VGGface_model} (trained on a large-scale dataset collected from the Internet). 
It trained the VGGNet on this dataset and  fine-tuned the networks via a triplet loss function, Similar to FaceNet. VGGface obtained a very high accuracy rate of 98.95\%.

In 2016, Liu and colleagues developed a "Large-Margin Softmax Loss" for CNNs \cite{Largesoftmax}, and showed its promise on multiple computer vision datasets, including LFW. They claimed that, cross-entropy does not explicitly encourage discriminative learning of features, and proposed a generalized large-margin softmax loss, which explicitly encourages intra-class compactness and inter-class separability between learned features.

In the same year, Wen et al. proposed a new supervision signal, called "center loss", for face recognition
task \cite{wen2016discriminative}. The center loss simultaneously learns a center for deep features of each class and penalizes the distances between the deep features and their corresponding class centers. With the joint supervision of softmax loss and center loss, they trained a CNN to obtain the deep features with the two key learning objectives, inter-class dispension and intra-class compactness as
much as possible.

In another work in 2016, Sun et al. proposed a face recognition model using a convolutional network with sparse neural connections \cite{sun2016sparsifying}.
This sparse ConvNet is learned in an iterative fashion, where each time one additional layer is sparsified and the entire model is re-trained given the initial weights learned in previous iterations (they found out training the sparse ConvNet from scratch usually fails to find good solutions for face recognition).

In 2017, in \cite{RangeLoss}, Zhang and colleagues developed a range loss to reduce the overall intra-personal variations while increasing inter-personal differences simultaneously.
%This model was trained on a large number of images from MS-Celeb-1M and CASIA-WebFace datasets, and achieved very promising results.
In the same year, Ranjan and colleagues developed an "L2-constraint softmax loss function" and used it for face verification \cite{L2softmax}. This loss function restricts the feature descriptors to lie on a hyper-sphere of a fixed radius. This work achieved  state-of-the-art performance on LFW dataset with an accuracy of 99.78\% at the time.
In \cite{cocoloss}, Liu and colleagues developed a face recognition model based on the intuition that the cosine distance of face features in high-dimensional space should be close enough within one class and far away across categories.
They proposed the congenerous cosine (COCO) algorithm to simultaneously optimize the cosine similarity among data.

In the same year, Liu et al. developed SphereFace \cite{Sphereface}, a deep hypersphere embedding  for face recognition.
They proposed an angular softmax (A-Softmax) loss function that enables CNNs to learn angular discriminative features.
Geometrically, A-Softmax loss can be viewed as imposing discriminative constraints on a hypersphere manifold, which intrinsically matches the prior that faces also lie on a manifold.
They showed promising face recognition accuracy on LFW, MegaFace, and Youtube Face databases.

In 2018, in \cite{amsloss} Wang et al. developed a simple and geometrically interpretable objective function, called additive margin Softmax (AM-Softmax), for deep face verification.
This work is heavily inspired by two previous works, Large-margin Softmax \cite{Largesoftmax}, and Angular Softmax in \cite{Sphereface}.
\iffalse
Through visualization they showed that large-margin Softmax can do a better job in learning face embeddings.
This is shown in Figure \ref{fig_embed}.
\begin{figure}[h]
\begin{center}
   \includegraphics[page=5,width=0.99\linewidth]{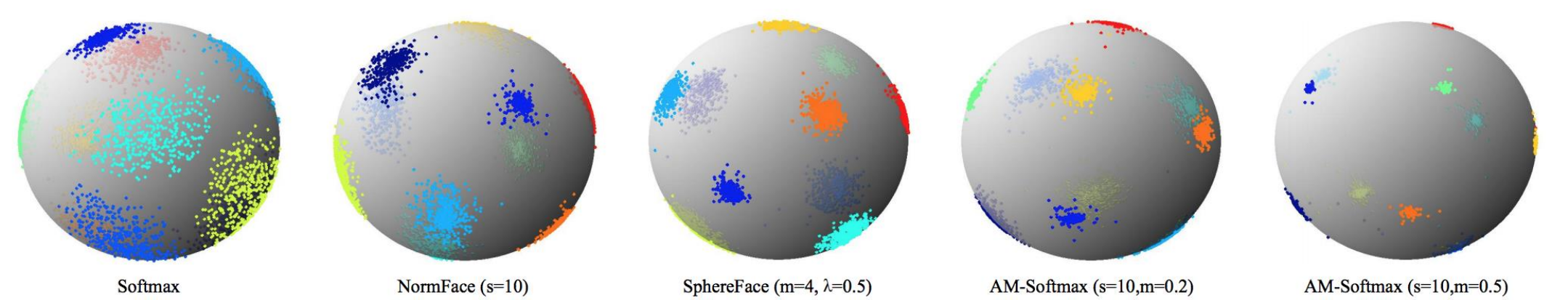}
\end{center}
   \caption{Feature distribution visualization of several loss functions. Each point on the sphere represents one normalized feature. Different
colors denote different classes. Courtesy of \cite{amsloss}.}
\label{fig_embed}
\end{figure}
\fi

CosFace \cite{CosFace} and ArcFace \cite{Arcface} are two other promising face recognition works developed in 2018.
In \cite{CosFace}, Wang et al. proposed a novel loss function, namely large margin cosine loss (LM-CL). 
More specifically, they reformulate the softmax loss as a cosine loss by L2 normalizing both features and
weight vectors to remove radial variations, based on which a cosine margin term is introduced to further maximize the decision margin in the angular space. 
As a result, minimum intra-class variance and maximum inter-class variance are achieved by virtue of normalization and cosine decision margin maximization.

\begin{figure}[h]
\begin{center}
   \includegraphics[page=20,width=0.7\linewidth]{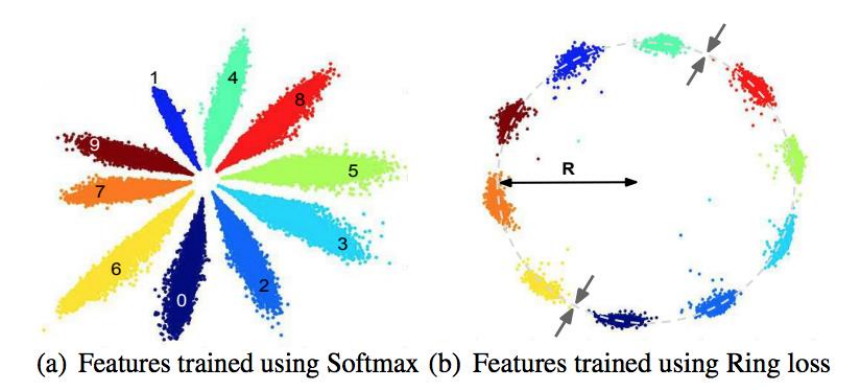}
\end{center}
   \caption{Sample MNIST features trained using (a) Softmax and (b) Ring loss on top of Softmax. Courtesy of \cite{RingLoss}.}
\label{ringloss}
\end{figure}

Ring-Loss \cite{RingLoss} is another work focused on designing a new loss function, which applies soft normalization, where it gradually learns to constrain the norm to the scaled unit circle while preserving convexity leading to more robust features.
The comparison of learned features by regular softmax and the Ring-loss function is shown in Figure \ref{ringloss}.
%As we can see many of the face recognition works have focused on developing new loss functions (such as center loss, large margin softmax loss, and angular softmax), which are more suitable for fine-grained discrimination. 

AdaCos \cite{adacos},  P2SGrad \cite{p2sgrad}, UniformFace \cite{uniface}, and AdaptiveFace \cite{adapface} are among the most promising works proposed in 2019. 
In AdaCos \cite{adacos}, Zhang et al. proposed a novel cosine-based softmax loss, AdaCos, which is hyperparameter-free and leverages an adaptive scale parameter to automatically strengthen the training supervisions during the training process.
In \cite{p2sgrad}, Zhang et al. claimed that cosine based losses always include sensitive
hyper-parameters which can make training process unstable, and it is very tricky to set suitable hyperparameters for a specific dataset.  
They addressed this challenge by directly designing the gradients for training in an adaptive manner.
P2SGrad was able to achieves state-of-the-art performance on all three face recognition benchmarks, LFW, MegaFace, and IJB-C.
%AdaptiveFace \cite{adapface}, and UniFace \cite{uniface} are two other promising work proposed in 2019.
There are several other works proposed for face recognition. For more detailed overview of deep learning-based face recognition, we refer the readers to \cite{deep_face_survey}.

\begin{figure}[h]
\begin{center}
   \includegraphics[page=26,width=0.7\linewidth]{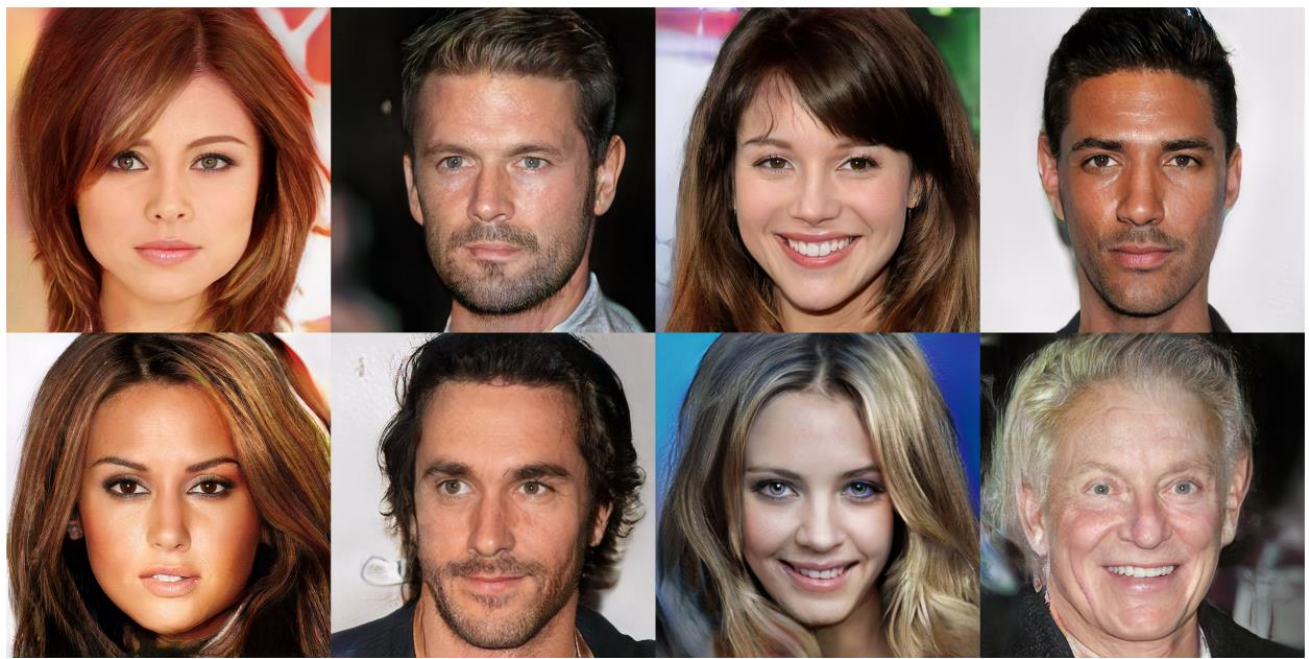}
\end{center}
   \caption{8 sample images (of 1024 x 1024) generated by progressive GAN, using the CELEBA-HQ dataset. Courtesy of \cite{prog_gan}.}
\label{fig:prog_gan}
\end{figure}

There have also been several works on using generative models for face image generation. 
To show the results of one promising model, in Progressive-GAN \cite{prog_gan}, Karras et al developed a framework to grow both the generator and discriminator of GAN progressively, which can learn to generate  high-resolution realistic images. Figure \ref{fig:prog_gan} shows 8 sample face images generated by this Progressive-GAN model trained on CELEB-A dataset.
.

\begin{figure*}[h]
\begin{center}
   \includegraphics[page=7,width=0.99\linewidth]{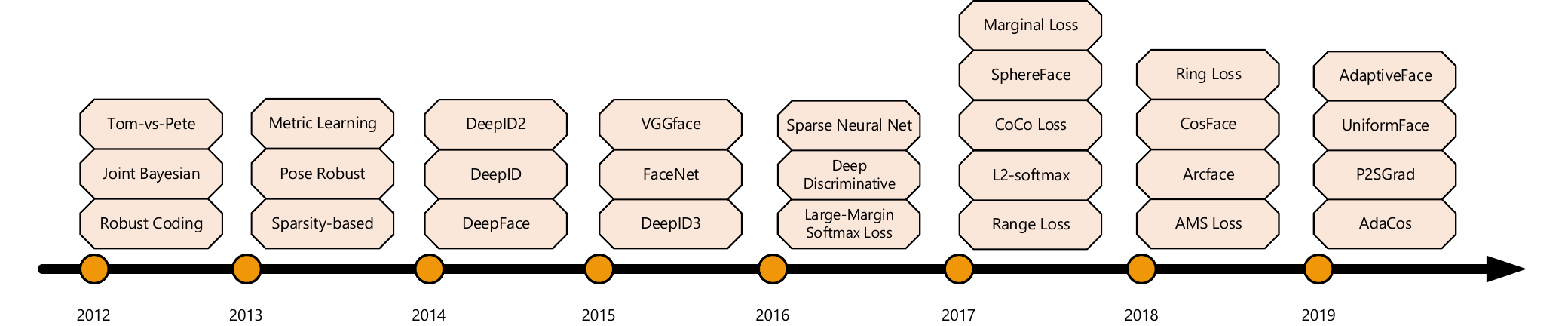}
\end{center}
\vspace{-3mm}
   \caption{A timeline of face recognition methods.}
\label{fig_face_timeline}
\end{figure*}

Figure \ref{fig_face_timeline} illustrates the timeline of popular face recognition models since 2012. 
The listed models after 2014 are all deep learning based models.
DeepFace and DeepID mark the beginning of deep learning based face recognition. 
As we can see many of the models after 2017 have focused on developing new loss functions for more discriminative feature learning.

\textbf{\subsection{Fingerprint Recognition}}
Fingerprint is arguably the most commonly used physiological biometric feature. It consists of ridges and valleys, which form unique patterns. Minutiae are major local portions of the fingerprint which can be used to determine the uniqueness of the fingerprint \cite{jain1997line}. Important features exist in a fingerprint include ridge endings, bifurcations, islands, bridges, crossovers, and dots. \cite{hrechak1990automated}.

A fingerprint needs to be captured by a special device in its close proximity. This makes making a dataset of fingerprints more time-consuming than some other biometrics, such as faces and ears. Nevertheless, there are quite a few remarkable fingerprint datasets that are being used around the world.
Fingerprint recognition has always been a very active area with wide applications in industry, such as smartphone authentication, border security, and forensic science. 
%In one of the earlier works in this area, \cite{hrechak1990automated} used structural matching for fingerprint recognition; eight main features would be identified in the print, and the list of features around each would be counted and used as a feature vector. Therefore, in this method, partial fingerprints could also be used. 
As one of the classical works, Lee et al \cite{lee1999fingerprint} used Gabor filtering on partitioned fingerprint images to extract features, followed by a k-NN classifier for the recognition, achieving 97.2\% recognition rate. 
In addition, using the magnitude of the filter output with eight orientations added a degree of shift-invariance to the recognition scheme. Tico et al \cite{tico2001wavelet} extracted wavelet features from the fingerprint to use in a k-NN classifier.

\textbf{\subsubsection{Fingerprint Datasets}}
There are several datasets developed for fingerprint recognition. Some of the most popular ones include:

\textbf{FVC Fingerprint Database:}
Fingerprint Verification Competition (FVC) is widely used for fingerprint evaluation \cite{fvc_finger}. 
FVC 2002 consists of three fingerprint datasets (DB1, DB2, and DB3) collected using different sensors. Each of these datasets consists of two sets: 
(i) Set A with
100 subjects and 8 impressions per subject, 
(ii) Set B with 10 subjects and 8 impressions per subject.
FVC 2004 adds another dataset (DB4) and contains more deliberate noise, e.g. skin distortions, skin moisture, and rotation.

\textbf{PolyU High-resolution Fingerprint Database:}
This dataset contains two high resolution fingerprint image databases (denoted as DBI and DBII), provided by the Hong Kong Polytechnic University \cite{polyU_finger}.
It contains 1480 images of 148 fingers.

\textbf{CASIA Fingerprint Dataset:}
CASIA Fingerprint Image Database V5 contains 20,000 fingerprint images of 500 subjects \cite{casia_finger}. 
%The fingerprint images of this dataset were captured using URU4000 fingerprint sensor in one session.
Each volunteer contributed 40 fingerprint images of his eight fingers (left and right thumb, second, third, fourth finger), i.e., 5 images per finger. 
The volunteers were asked to rotate their fingers with various levels of pressure to generate significant intra-class variations. 
%All fingerprint images are 8 bit gray-level BMP files and the image resolution is 328*356.

\textbf{NIST Fingerprint Dataset:}
NIST SD27 consists of 258 latent fingerprints and corresponding reference fingerprints \cite{nist_finger}.

\textbf{\subsubsection{Deep Learning Works on Fingerprint Recognition}}
There have been  numerous works on using deep learning for fingerprint recognition. 
Here we provide a summary of some of the prominent works in this area.

\begin{figure*}[h]
\begin{center}
   \includegraphics[page=7,width=0.8\linewidth]{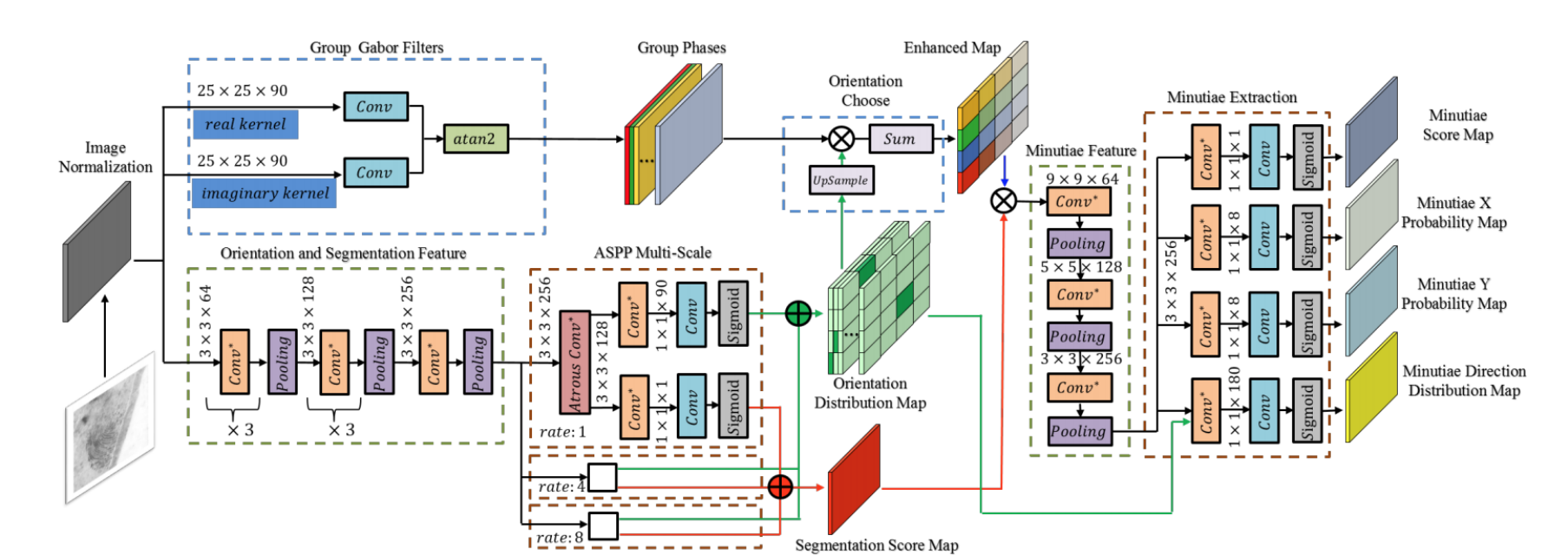}
\end{center}
   \caption{The block-diagram of the proposed FingerNet model for minutiae extraction. Courtesy of \cite{finger_2}.}
\label{fig:fingernet_arch}
\end{figure*}

In \cite{finger_1}, Darlow et al. proposed a fingerprint minutiae extraction algorithm based on deep learning models, called MENet, and achieve promising results on fingerprint images from FVC datasets.
In \cite{finger_2}, Tang and colleagues proposed another deep learning-based model for fingerprint minutiae extraction, called FingerNet. This model  jointly performs feature extraction, orientation estimation, segmentation, and uses them to estimate the minutiae maps. 
The block-diagram of this model is shown in Figure \ref{fig:fingernet_arch}.
.

In another work \cite{finger_3}, Lin and Kumar proposed a multi-view deep representation (based on CNNs) for contact-less and partial 3D fingerprint recognition. The proposed model includes one fully convolutional network for fingerprint segmentation and three Siamese networks to learn multi-view 3D fingerprint feature representation.
They show promising results on several 3D fingerprint databases.
In \cite{finger_4}, the authors develop a fingerprint texture learning using a deep learning framework.
They evaluate their models on several benchmarks, and achieve verification accuracies of 100, 98.65, 100 and 98\% on the four databases of PolyU2D, IITD, CASIA-BLU and CASIA-WHT, respectively. 
In \cite{finger_5}, Minaee et al. proposed a deep transfer learning approach to perform fingerprint recognition with a very high accuracy. They fine-tuned a pre-trained ResNet model on a popular fingerprint dataset, and are able to achieve very high recognition rate. 

In \cite{finger_8}, Lin and Kumar proposed a multi-Siamese network to accurately match contactless to contact-based fingerprint images. In addition to the fingerprint images, hand-crafted fingerprint features, e.g., minutiae and core point, are also incorporated into the proposed architecture. This multi-Siamese CNN is trained using the fingerprint images and extracted features.
%This model achieved promising results over other CNN-based methods and the traditional fingerprint cross matching methods.

There are also some works using deep learning models for fingerprint segmentation.
In \cite{finger_6}, Stojanovic and colleagues proposed a fingerprint ROI segmentation algorithm based on convolutional neural networks. 
In another work \cite{finger_7}, Zhu et al. proposed a new latent fingerprint segmentation method based on convolutional neural networks ("ConvNets").
The latent fingerprint segmentation problem is formulated as a classification system, in which a set of ConvNets are trained to classify each patch as either fingerprint or background. 
%Considering the spatial correlation between fingerprint patches, they trained the set of ConveNets using multi-sized overlapping patches to utilize complementary information. 
Then, a score map is calculated based on the classification results to evaluate the possibility of a pixel belonging to the fingerprint foreground. Finally, a segmentation mask is generated by thresholding the score map and used to delineate the latent fingerprint boundary.
%They evaluate their results on NIST SD27 latent database, and achieved promising results in terms of both false detection rate (FDR) and overall segmentation accuracy.

There have also been some works for fake fingerprint detection.
In \cite{finger_9}, Kim et al. proposed a fingerprint liveliness detection based on statistical features learned from deep belief network (DBN).
This method achieves good accuracy on various sensor datasets of the LivDet2013 test.
In \cite{finger_10}, Nogueira and colleagues proposed a model to detect fingerprint liveliness (where they are real or fake) using a convolutional neural network, which achieved an accuracy of 95.5\% on fingerprint liveness detection competition 2015.

There have also been some works on using generative models for fingerprint image generation. 
In \cite{finger_11}, Minaee et al proposed an algorithm for fingerprint image generation based on an extension of GAN, called "Connectivity Imposed GAN". This model adds total variation of the generated image to the GAN loss function, to promote the connectivity of generated fingerprint images. 
In \cite{anil_finger}, Tabassi et al. developed a framework to synthesize altered fingerprints whose characteristics are similar to true altered fingerprints, and used them to train a classifier to detect "Fingerprint alteration/obfuscation
presentation attack" (i.e. intentional tamper or damage to the real friction ridge patterns to avoid identification).

\iffalse
Four sample fingerprint images generated by this work (over different training epochs) are shown in \ref{fig:finger_gan1}. As we can see the generated samples get more and more realistic as training iteration proceeds.
\begin{figure}[h]
\begin{center}
   \includegraphics[page=6,width=0.95\linewidth]{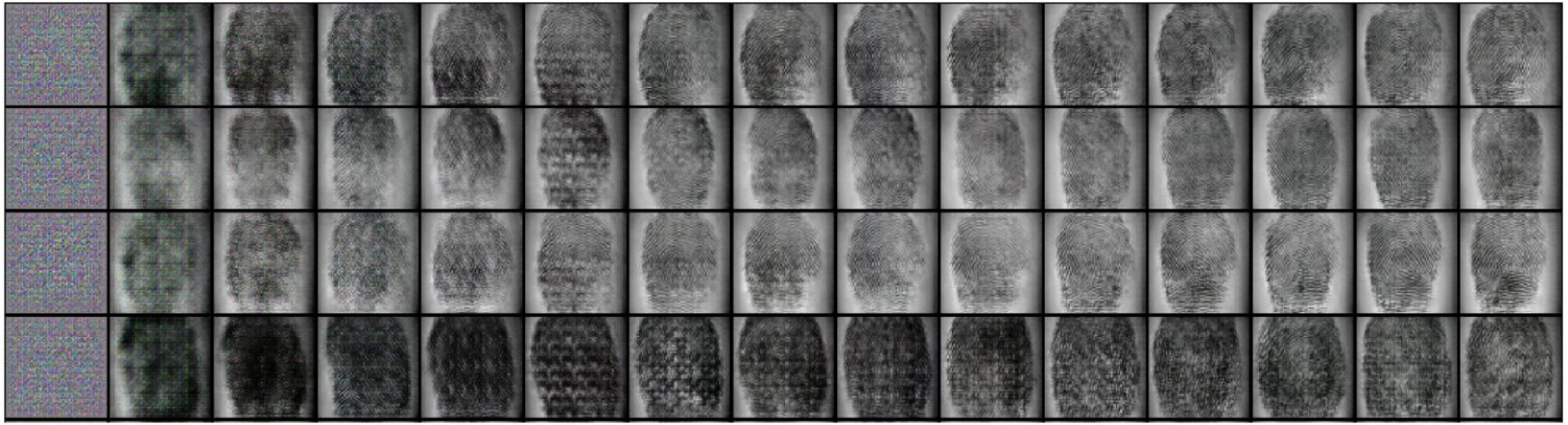}
\end{center}
   \caption{ The generated fingerprint images for 4 input latent vectors, over 140 epochs (on every 10 epochs), using the trained model on FVC-2006 fingerprint database.}
\label{fig:finger_gan1}
\end{figure}
\fi

\textbf{\subsection{Iris Recognition}}
Iris images contain a rich set of features embedded in their texture and patterns which do not change over time, such as rings, corona, ciliary processes, freckles, and the striated trabecular meshwork of chromatophore and fibroblast cells, which is the most prevailing under visible light \cite{daugman1993high}. Iris recognition has gained a lot of attention in recent years in different security-related fields. 

\begin{figure*}[h]
\begin{center}
   \includegraphics[page=10,width=0.35\linewidth]{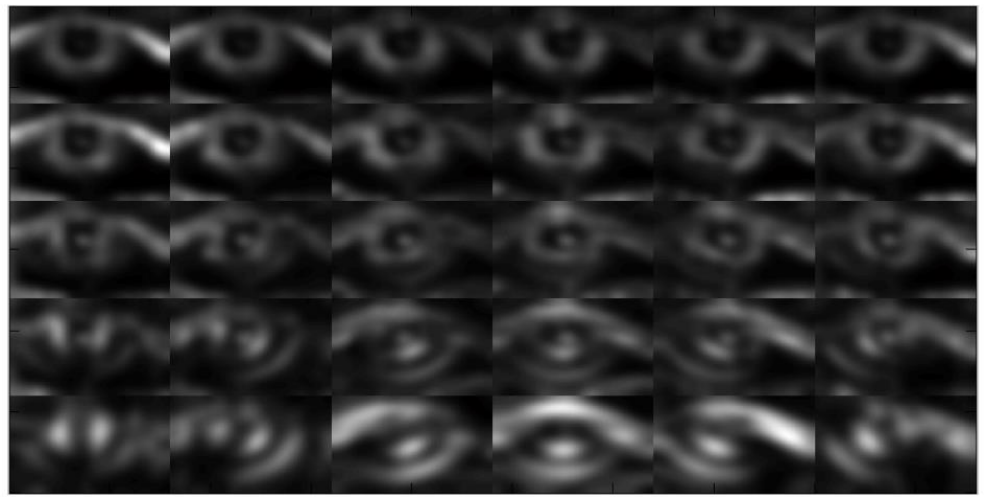}
   \includegraphics[page=11,width=0.35\linewidth]{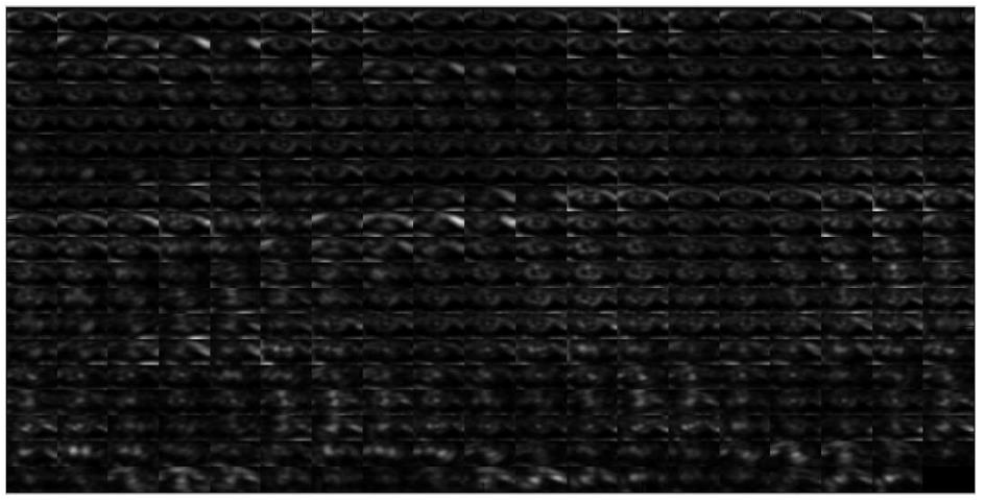}
\end{center}
   \caption{The images from the first (on the left) and second (on the right) layers of the scattering transform.}
\label{fig:iris_scat1}
\end{figure*}

John Daugman developed one of the first modern iris recognition frameworks using 2D Gabor wavelet transform \cite{kumar2010comparison}.
Iris recognition started to rise in popularity in the 1990s. 
In 1994, Wildes et al \cite{wildes1994system} introduced a device using iris recognition for personnel authentication. After that, many researchers started looking at iris recognition problem.
Early works have used a variety of methods to extract hand-crafted features from the iris. Williams et al \cite{williams1996iris} converted all iris entries to an ``IrisCode'' and used Hamming's distance of an input iris image's IrisCode from those of the irises in the database as a metric for recognition. 
%\cite{boles1998human} used zero-crossings of wavelet transform extracted from a collection of circular contours extracted from the iris image.
%\cite{zhu2000biometric} used multi-channel Gabor filtering and wavelet transform for iris identification.
In \cite{iris_0}, the authors proposed an iris recognition system based on "deep scattering convolutional features", which achieved a significantly high accuracy rate on IIT Delhi dataset.
This work is not exactly using deep learning, but is using a deep scattering convolutional network, to extract hierarchical features from the image.
The output images at different nodes of scattering network denote the transformed image along different orientation and scales. 
The transformed images of the first and second layers of scattering transform for a sample iris image are shown in Figures \ref{fig:iris_scat1}. 
These images are derived by applying bank of filters of 5 different scales and 6 orientations.

\iffalse
Deep scattering network is a multi-layer network with pre-defined weights based on wavelet analysis. It provides a higher-order wavelet decomposition of an image along different scales and orientations, and is shown to capture higher order statistics \cite{scat1}.
The block diagram of a scattering network with 3 layers is shown in Figure \ref{fig:scat_net}.
\begin{figure}[h]
\begin{center}
   \includegraphics[page=1,width=0.6\linewidth]{img/scatnet.png}
\end{center}
  \caption{Scattering operator applied up to three layers. The output signals in each layer are used as the input for the scattering transformation in the next layer of scattering network [?]. The scattering operator at each layer is simply the cascade of three operations: wavelet decomposition, complex modulus (phase removal) and a local averaging.}
\label{fig:scat_net}
\end{figure}
\fi

\begin{figure}[h]
\begin{center}
   \includegraphics[page=9,width=0.8\linewidth]{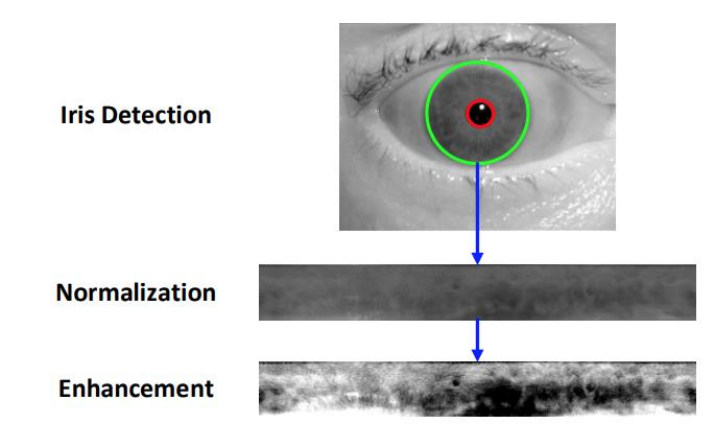}
\end{center}
   \caption{Illustration of some of the key pre-processing steps for iris recognition, courtesy of \cite{iris_12}.}
\label{fig:iris_preprocess}
\end{figure}

It is worth mentioning that many of the classical iris recognition models perform several pre-processing steps such as iris detection, normalization, and enhancement, as shown in Figure  \ref{fig:iris_preprocess}. They then extract features from the normalized or enhanced image. Many of the modern works on iris recognition skip normalization and enhancement, and yet, they are still able to achieve very high recognition accuracy. 
One reason is the ability of deep models to capture high-level semantic a features from original iris images, which are discriminative enough to perform well for iris recognition.

\textbf{\subsubsection{Iris Datasets}}
Various datasets have been proposed for iris recognition in the past. Some of the most popular ones include:

\textbf{CASIA-Iris-1000 Database:}
CASIA-Iris-1000 contains 20,000 iris images from 1,000 subjects, which were collected using an IKEMB-100 camera. The main sources of intra-class variations in CASIA-Iris-1000 are eyeglasses and specular reflections \cite{casia_iris}. 
%The IIT Delhi database contains 2240 iris images captured from 224 different people. The resolution of these images is 320x240 pixels.

\textbf{UBIRIS Dataset:}
The UBIRIS database has two distinct versions, UBIRIS.v1 and UBIRIS.v2. 
The first version of this database is composed of 1877 images collected from 241 eyes in two distinct sessions. It simulates less constrained imaging conditions \cite{ubiris_iris}.
The second version of the UBIRIS database has over 11000 images (and continuously growing) and more realistic noise factors.

\textbf{IIT Delhi Iris Dataset:}
IIT Delhi iris database contains 2240 iris images captured from 224 different people. 
The resolution of these images is 320x240 pixels \cite{iit_iris}.
%Six sample images from this dataset are shown in Fig 4. As we can see the
Iris images in this dataset have variable color distribution, and different (iris) sizes.

\textbf{ND Datasets:}
ND-CrossSensor-Iris-2013 consists of two iris databases, taken with two iris sensors: LG2200 and LG4000. The LG2200 dataset consists of 116,564 iris images, and LG4000 consists of 29,986 iris images of 676 subjects \cite{lg_iris}.

\textbf{MICHE Dataset:}
Mobile Iris Challenge Evaluation (MICHE) consists of iris images acquired under unconstrained conditions using smartphones. It consists of more than 3,732 images acquired from 92 subjects using three different smartphones \cite{miche_iris}.

\textbf{\subsubsection{Deep Learning Works on Iris Recognition}}
Compared to face recognition, deep learning models made their ways to iris recognition with a few years delay.

As one of the first works using deep learning for iris recognition, in \cite{iris_1} Minaee et al. showed that features extracted from a pre-trained CNN model trained on ImageNet are able to achieve a reasonably high accuracy rate for iris recognition.
In this work, they used  features derived from different layers of VGGNet \cite{vggnet}, and trained a multi-class SVM on top of it, and showed that the trained model can achieve state-of-the-art accuracy on two iris recognition benchmarks, CASIA-1000 and IIT Delhi databases. 
They also showed that features extracted from the mid-layers of VGGNet achieve slightly higher accuracy from the the very last layers.
In another work \cite{iris_1_2}, Gangwar and Joshi proposed an iris recognition network based on convolutional neural network, which provides robust, discriminative, compact resulting in very high accuracy rate, and can work pretty well in cross-sensor recognition of iris images.

In \cite{iris_1_3}, Baqar and colleagues proposed an iris recognition framework based on  deep belief networks, as well as contour information of iris images. Contour based feature vector has been used to discriminate samples belonging to different classes i.e., difference of sclera-iris and iris-pupil contours, and is named as “Unique Signature”. 
%Contours of both the boundaries are extracted using the radius vector function. 
Once the features extracted, deep belief network (DBN) with modified back-propagation algorithm based feed-forward neural network (RVLR-NN) has been used for classification. 
%They evaluate the proposed model on CASIA iris databases, and report the accuracy for different level of noise added to the images, and were able to achieve reasonably high accuracy rate, even in noisy image setting.

\begin{figure}[h]
\begin{center}
   \includegraphics[page=12,width=0.7\linewidth]{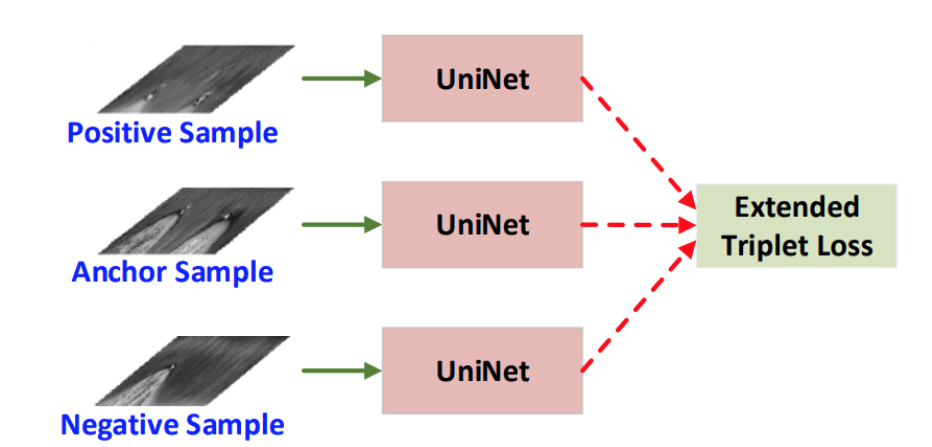}
\end{center}
   \caption{The block-diagram of triplet network used for iris recognition, courtesy of \cite{iris_12}.}
\label{fig:iris_triplet}
\end{figure}

In \cite{iris_12}, Zhao and Kumar proposed an iris recognition model based on "Deeply Learned Spatially Corresponding Features".
The proposed framework is based on a fully convolutional network (FCN), which outputs spatially corresponding iris feature descriptors. 
They also introduce a specially designed "Extended Triplet Loss (ETL)" function to incorporate the bit-shifting and non-iris masking. 
The triplet network is illustrated in Figure \ref{fig:iris_triplet}. 
They also developed a sub-network to provide appropriate information for identifying meaningful iris regions, which serves as essential input for the newly developed ETL.
They were able to outperform several classic and state-of-the-art iris recognition approaches on a few iris databases.

In another work \cite{iris_2}, Alaslani et al. developed an iris recognition system, based on deep features extracted from AlexNet, followed by a multi-class classification, and achieved high accuracy rates on CASIA-Iris-V1, CASIA-Iris-1000 and, CASIA-Iris-V3 Interval databases.
In \cite{menon2018iris}, Menon shows the applications of convolutional features from a fine-tuned pre-trained model for both identification and verification problems. 
In \cite{iris_4}, Hofbauer and colleagues proposed a CNN based algorithm for segmentation of iris images, which can results in higher accuracies than previous models.
In another work \cite{iris_5}, Ahmad and Fuller developed an iris recognition model based on triplet network, call ThirdEye.
Their work directly uses the segmented, un-normalized iris images, and is shown to achieve equal error rates of 1.32\%, 9.20\%, and 0.59\% on the ND-0405, UbirisV2, and IITD datasets respectively.
In a more recent work \cite{iris_8}, Minaee and colleagues proposed an algorithm for iris recognition based using a deep transfer learning approach. They trained a CNN model (by fine-tuning a pre-trained ResNet model) on an iris dataset, and achieved very accurate recognition on the test set.

\begin{figure}[h]
\begin{center}
   \includegraphics[page=8,width=0.999\linewidth]{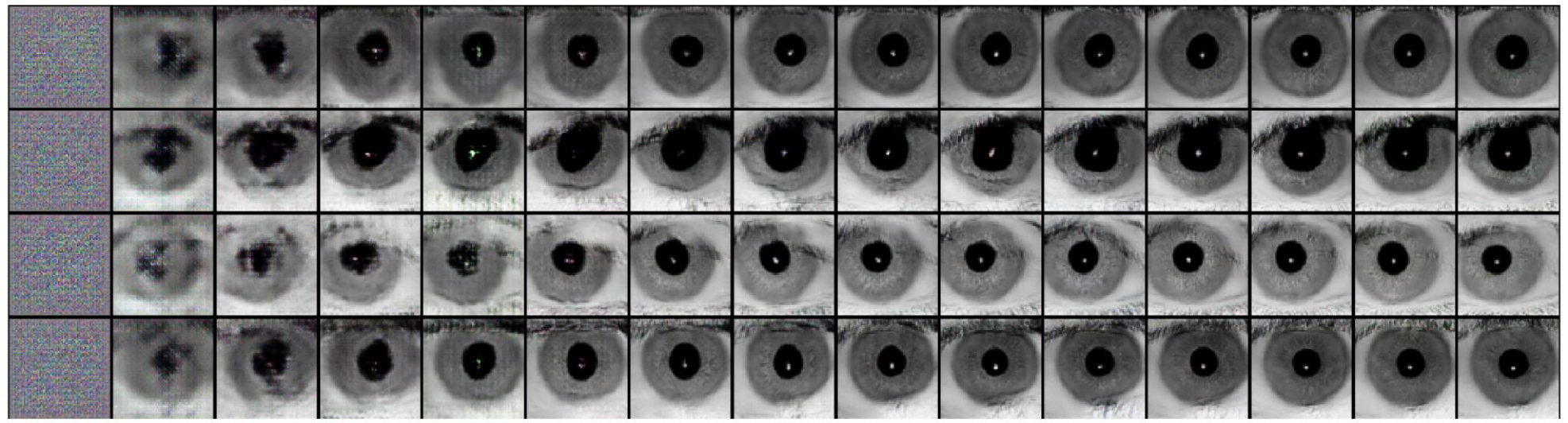}
\end{center}
   \caption{ The generated iris images for 4 input latent vectors, over 140 epochs (on every 10 epochs), using the trained model on IIT Delhi Iris database. Courtesy of \cite{iris_6}.}
\label{fig:iris_gan1}
\end{figure}

With the rise of deep generative models, there have been works that apply them to iris recognition. 
In \cite{iris_6}, Minaee et al proposed an algorithm for iris image generation based on convolutional GAN, which can generate realistic iris images. These images can be used for augmenting the training set, resulting in better feature representation and higher accuracy. 
Four sample iris images generated by this work (over different training epochs) are shown in Figure \ref{fig:iris_gan1}. 
%As we can see the generated samples get more and more realistic as training iteration proceeds.

In \cite{iris_7}, Lee and colleagues proposed a data augmentation technique based on GAN to augment the training data for iris recognition, resulting in a higher accuracy rate.
They claim that historical data augmentation techniques such as geometric transformations and brightness adjustment result in samples with very high correlation with the original ones, but using augmentation based on a conditional generative adversarial network can result in higher test accuracy.

\textbf{\subsection{Palmprint Recognition}}
Palmprint is another biometric which is gaining more attention recently. In addition to minutiae features, palmprints also consist of geometry-based features, delta points, principal lines, and wrinkles \cite{zhang1999two, zhang2012comparative}.
Each part of a palmprint has different features, including texture, ridges, lines and creases. 
An advantage of palmprints is that the creases in palmprint virtually do not change over time and are easy to extract \cite{chen2001palmprint}. 
However, sampling palmprints requires special devices, making their collection not as easy as other biometrics such as fingerprint, iris and face. %Palmprint recognition seems to have gained more attention starting from the late 1990s and early 2000s, when handcrafted features and techniques were used, before the popularity of deep learning-based methods. 
Classical works on palmprint recognition have explored a wide range of hand-carfted features such as as PCA and ICA \cite{connie2003palmprint}, Fourier transform \cite{li2002palmprint}, wavelet transform \cite{wu2002wavelet}, line feature matching \cite{shu1998palmprint}, and deep scaterring features \cite{palm_2}.

\textbf{\subsubsection{Palmprint Datasets}}
Several datasets have been proposed for palmprint recognition dataset. Some of the most widely used datasets include:

\textbf{PolyU Multispectral Palmprint Dataset:}
The images from PolyU dataset were collected from 250 volunteers, including 195 males and 55 females. 
In total, the database contains 6,000 images from 500 different palms for one illumination \cite{polyu_palm}.
Samples are collected in two separate sessions. 
In each session, the subject was asked to provide 6 images for each palm. 
Therefore, 24 images of each illumination from 2 palms were collected from each subject.

\textbf{CASIA Palmprint Database:}
CASIA Palmprint Image Database contains of 5,502 palmprint images captured from 312 subjects. 
For each subject, they collect palmprint images from both left and right palms \cite{casia_palm}. 
All palmprint images are 8-bit gray-level JPEG files by their self-developed palmprint recognition device.

\textbf{IIT Delhi Touchless Palmprint Database:}
The IIT Delhi palmprint image database consists of the hand images collected from the students and staff at IIT Delhi, New Delhi, India \cite{iit_palm}. This database has been acquired using a simple and touchless imaging setup. The currently available database is from 235 users. 
Seven images from each subject, from each of the left and right hand, are acquired in varying hand pose variations. Each image has a size of 800x600 pixels. 
%In addition to the original images, 150x150 pixel automatically cropped and normalized palmprint images are also available.

\textbf{\subsubsection{Deep Learning Works on Palmprint Recognition}}
In \cite{palm_1}, Xin et al. proposed one of the early works on palmprint recognition using a deep learning framework.
The authors built a deep belief net by top-to-down unsupervised training, and tuned the model parameters toward a robust accuracy on the validation set.
Their experimental analysis showed a performance gain over classical models that are based on LBP, and PCA, and other other hand-crafted features.

\iffalse
In another work \cite{palm_2}, Minaee and Wang proposed a palmprint recognition using deep scattering features, which are extracted from scattering convolutional network. This features are known to provide higher order statistical information which are missing in simple features such as SIFT, HoG, etc. The proposed framework has been tested on a well-known PolyU multispectral palmprint database, and achieved accuracy rates of 99.4\% and 99.9\%, which beat the previous state-of-the-art performance. 
The transformed images from the 2nd layer of scattering network for a sample palmprint images are shown in Figure
\ref{fig:palm_scat2}.
\begin{figure}[h]
\begin{center}
   \includegraphics[page=1,width=0.6\linewidth]{img/palm_scat2.png}
\end{center}
   \caption{ The transformed images from the 2nd layer of scattering convolutional network.}
\label{fig:palm_scat2}
\end{figure}
\fi

In another work, Samai et al. proposed a deep learning-based model for 2D and 3D palmprint recognition \cite{palm_3}. 
They proposed an efficient biometric identification system combining 2D and 3D palmprint by fusing them at matching score level. 
To exploit the 3D palmprint data, they converted them to grayscale images by using the Mean Curvature (MC) and the Gauss Curvature (GC). They then extracted features from images using Discrete Cosine Transform Net (DCT Net).

Zhong et al. proposed a palmprint recognition algorithm using Siamese network \cite{palm_4}.
Two VGG-16 networks (with shared parameters) were employed to extract features for two input palmprint images, and another network is used on top of them to directly obtain the similarity of two input palmprints according to their convolutional features. 
This method achieved  an Equal Error Rate (EER) of 0.2819\% on on PolyU dataset. 
In \cite{palm_5}, Izadpanahkakhk et al. proposed a transfer learning approach towards palmprint verification, which jointly extracts regions of interests and features from the images.
They use a pre-trained convolutional network, along with SVM to make prediction. They achieved an IoU score of 93\% and EER of 0.0125 on Hong Kong Polytechnic University Palmprint (HKPU) database.

\begin{figure}[h]
\begin{center}
   \includegraphics[page=3,width=0.99\linewidth]{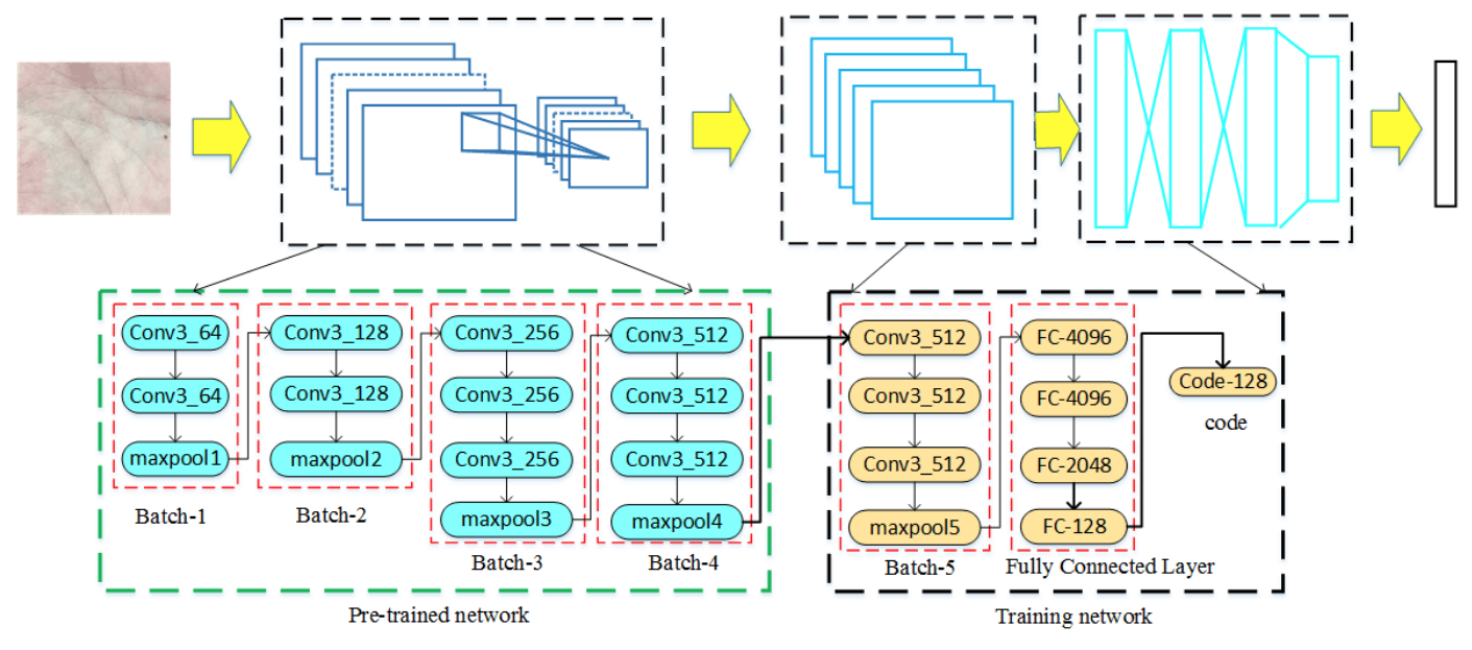}
\end{center}
   \caption{The block-diagram of the proposed deep hashing network, courtesy of \cite{palm_7}.}
\label{fig:DHN}
\end{figure}

In \cite{palm_6}, Shao and Zhong proposed a few-shot palmprint recognition model using a graph neural network.
In this work, the palmprint features extracted by a convolutional neural network are processed into nodes in the GNN. The edges in the GNN are used to represent similarities between image nodes. 
%The experimental results show that their proposed GNN-based few-shot palmprint recognition can obtain state-of-the-art performance, where the accuracy is over 99.90%.
In a more recent work \cite{palm_7}, Shao and colleagues proposed a deep palmprint recognition approach by combining hash coding and knowledge distillation.
Deep hashing network are used to convert palmprint images to binary codes to save storage space and speed up the matching process. 
The architecture of the proposed deep hashing network is shown in Figure \ref{fig:DHN}.
They also proposed a database for unconstrained palmprint recognition, which consists of more than 30,000 images collected by 5 different mobile phones, and achieved promising results on that dataset. 
In \cite{palm_8}, Shao et al. proposed a cross-domain palmprint recognition based on transfer convolutional autoencoder. Convolutional autoencoders were firstly used to extract low-dimensional features. A discriminator was then introduced to reduce the gap of two domains. The auto-encoders and discriminator were alternately trained, and finally the features with the same distribution were extracted.

In \cite{palm_9}, Zhao and colleagues proposed a joint deep convolutional feature representation for hyperspectral palmprint recognition. A CNN stack is constructed to extract its features from the entire spectral bands and generate a joint convolutional feature.
They evaluated their model on a hyperspectral palmprint dataset consisting of 53 spectral bands with 110,770 images.
They achieved an EER of 0.01\%.
In \cite{palm_10}, Xie et al. proposed a gender classification framework using convolutional neural network on plamprint images. They fine-tuned the pre-trained VGGNet on a palmprint dataset and showed that the proposed structure could achieve a good performance for gender classification.

\textbf{\subsection{Ear Recognition}}
Ear recognition is a more recent problem that scientists are exploring, and the volume of biometric recognition works involving ears is expected to increase in the coming years. One of the more prominent aspects of ear recognition is the fact that the subject can be photographed from either side of their head and the ears are almost identical (suitable when subject is not cooperating, or hiding his/her face). 
Also, since there is no need for the subject's proximity, images may be taken from the ear more easily. However, ears of the subject may still be occluded by factors such as hair, hat, and jewelry,  making it difficult to detect and use the ear image \cite{cintas2019automatic}. 
%Issues complicating the ear imaging may also include the camera angle and the weather.
There are multiple classical methods to perform ear recognition: geometric methods, which try to extract the shape of the ear; holistic methods, which extract the features from the ear image as a whole; local methods, which specifically use a portion of the image; and hybrid methods, which use a combination of the others \cite{emervsivc2017training}, \cite{naseem2008sparse}.

\textbf{\subsubsection{Ear Datasets}}
%Compared to the above biometrics, there are fewer datasets developed for ear recognition. 
The datasets below are some of the popular 2D ear recognition datasets, which are used by researchers.

\textbf{IIT Ear Database:}
The IIT Delhi ear image database contains 471 images,  acquired from the 121 different subjects and each subject has at least three ear images \cite{iit_ear}. 
All the subjects in the database are in the age group 14-58 years. 
The resolution of these images is 272x204 pixels and all these images are available in jpeg format. 
%In addition to the original images, this database also provide the automatically normalized and cropped ear images of size 50x180 pixels. 

\textbf{AWE Ear Dataset:}
This database contains 1,000 images of 100 persons. Images were collected from the web using a semi-automatic procedure, and contain the following annotations: gender, ethnicity, accessories, occlusions, head pitch, head roll, head yaw, head side, and central tragus point \cite{awe_ear}.

\textbf{Multi-PIE Ear Dataset:}
This dataset was created in 2017 \cite{eyiokur2017domain} based on the Multi-PIE face dataset \cite{gross2010multi}. There are 17,000 ear images extracted from the profile and near-profile images of 205 subjects present in the face dataset. The ears in the images are in different illuminations, angles, and conditions, making it a decent dataset for a more generalized ear recognition approach.

\textbf{USTB Ear Database:}
This dataset contains ear images of 60 volunteers captured in 2002 \cite{ustb_ear}.
Every volunteer is photographed three different images. They are normal frontal image, frontal image with trivial angle rotation and image under different lighting condition. 
%Each of them has 256 gray scales. Images had already experienced rotation and shearing, but they were without illumination compensation. 

\textbf{UERC Ear Dataset:}
The ear images in this dataset \cite{emervsivc2017unconstrained} are collected from the Internet in unconstrained conditions, i.e., from the wild. 
There is a total of 11,804 images from 3,706 subjects, of which 2,304 images from 166 subjects are for training, and the rest are for testing.

\textbf{AMI Ear Dataset:}
This dataset \cite{gonzalez2008biometria} contains 700 images of size 492 x 702 from 100 subjects in the age range of 19 to 65 years old. The images are all in the same lighting condition and distance, and from both sides of the subject's head. The images, however, differ in focal lengths, and the direction the subject is looking (up, down, left, right).

\textbf{CP Ear Dataset:}
One of the older datasets in this area, the Carreira-Perpinan dataset \cite{perpinan1995compression} contains 102 left ear images taken from 17 subjects in the same conditions.

\textbf{WPUT Ear Dataset:}
The West Pomeranian University of Technology (WPUT) dataset \cite{frejlichowski2010west} contains 2,071 images from 501 subjects (247 male and 254 female subjects), from different age groups and ethnicities. The images are taken in different lighting conditions, from various distances and two angles, and include ears with and without accessories, including earrings, glasses, scarves, and hearing aids.

\textbf{\subsubsection{Deep Learning Works on Ear Recognition}}
Ear recognition is not as popular as face, iris, and fingerprint recognition yet. Therefore, datasets used for this procedure are still limited in size. 
Based on this, Zhang et al \cite{zhang2019few} proposed few-shot learning methods, where the network use the limited training and quickly learn to recognize the images. Dodge et al \cite{dodge2018unconstrained}, who proposed using transfer learning with deep networks for unconstrained ear recognition Emersic and colleagues \cite{emervsivc2017unconstrained}, also proposed a deep learning-based averaging system to mitigate the overfitting caused by the small size of the datasets.
In \cite{emersic2017training}, the authors proposed the first publicly available CNN-based ear recognition method. 
They explored different strategies, such as different architectures, selective learning on pre-trained data and aggressive data augmentation to find the best configurations for their work. 

In \cite{emervsivc2018towards}, the authors showed how ear accessories can disrupt the recognition process and even be used for spoofing, especially in a CNN-based method, e.g., VGG-16, against a traditional method, e.g., local binary patterns (LBP), and proposed methods to remove such accessories and improve the performance, such as "inprinting" and area coloring.
Sinha et al \cite{sinha2019convolutional} proposed a framework which localizes the outer ear image using HOG and SVMs, and then uses CNNs to perform ear recognition. It aims to resolve the issues usually associated with feature extraction appearance-based techniques, namely the conditions in which the image was taken, such as illumination, angle, contrast, and scale, which are also present in other biometric recognition systems, e.g. for face.
Omara et al \cite{omara2018learning} proposed extracting hierarchical deep features from ear images, fusing the features using discriminant correlation analysis (DCA) Haghighat et al \cite{haghighat2016discriminant} to reduce their dimensions, and due to the lack of ear images per person, creating pairwise samples and using pairwise SVM \cite{brunner2012pairwise} to perform the matching (since regular SVM would not perform well due to the small size of the datasets). 
\iffalse
The network used is shown in Figure \ref{fig:omara2018learning}.
\begin{figure*}[h]
\begin{center}
   \includegraphics[page=18,width=0.8\linewidth]{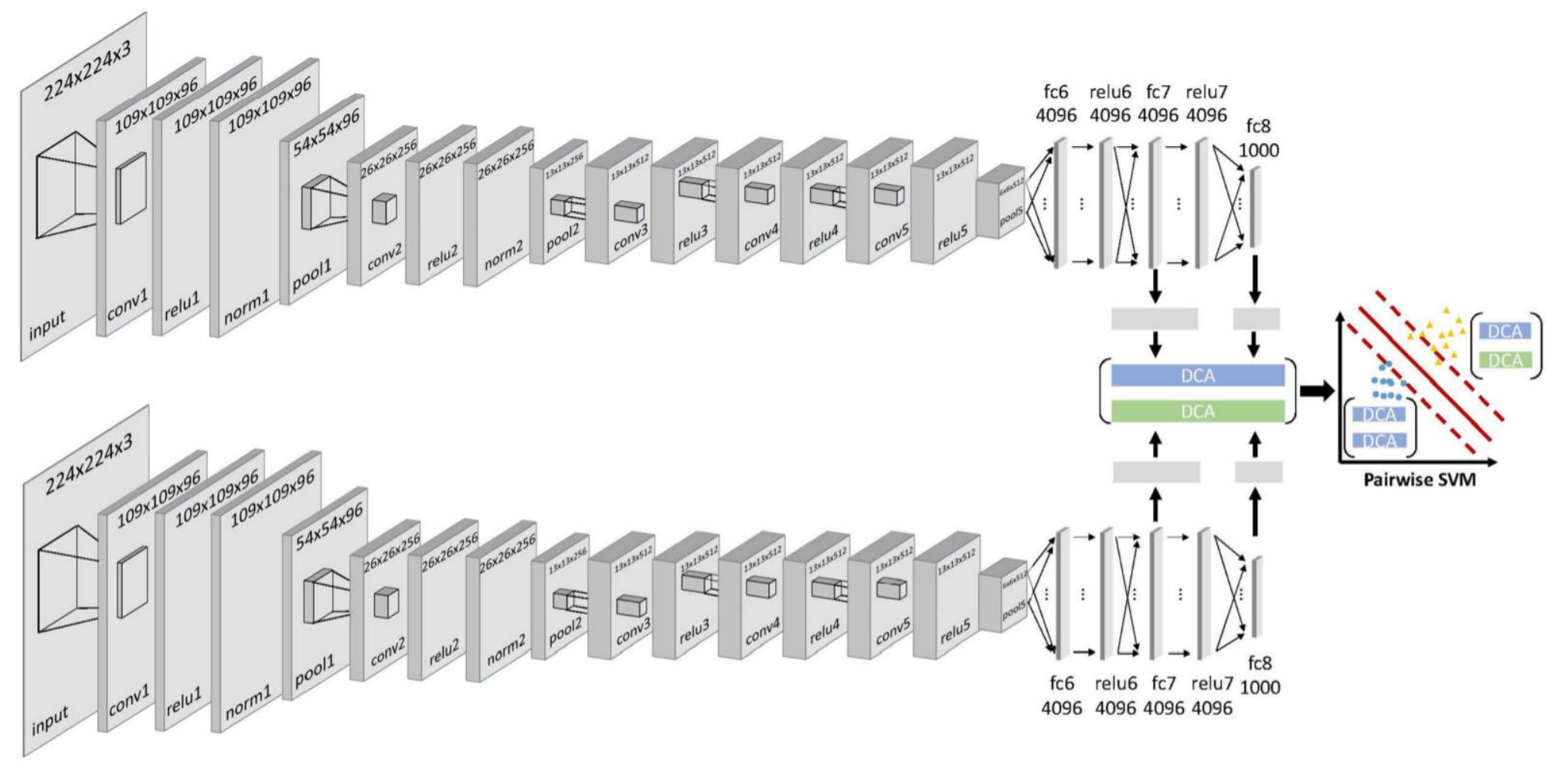}
\end{center}
   \caption{Ear recognition using hierarchical deep features and pairwise SVM, courtesy of \cite{omara2018learning}.}
\label{fig:omara2018learning}
\end{figure*}
\fi
Hansley et al \cite{hansley2018employing} used a fusion of CNNs and handcrafted features for ear recognition which outperformed other state-of-the-art CNN-based works, reaching to the conclusion that handcrafted features can complement deep learning methods.

\subsection{Voice Recognition}
Voice Recognition (also known as speaker recognition) is the task of determining a person's ID using the characteristics of one's voice. 
In a way, speaker recognition includes both behavioral and physiological features, such as accent and pitch respectively.
Using automatic ways to perform  speaker recognition dates back to 1960s when Bell Laboratories were approached by law enforcement agencies about the possibility of identifying callers who had made verbal bomb threats over the telephone \cite{shaver2016brief}. Over the years, researchers have  developed many models that can perform this task effectively, especially with the help of deep learning.
In addition to security applications, it is also being used in virtual personal assistants, such as Google Assistant, so they can recognize and distinguish the phone owner's voice from the others \cite{yoffie2018voice}.

Speaker recognition can be classified into speaker identification and speaker verification. 
speaker identification is the process of determining a person's ID from a set of registered voice using a given utterance \cite{naseem2010sparse}, whereas speaker verification is the process of accepting or rejecting a proposed identity claimed for a speaker \cite{Furui2008}. 
Since these two tasks usually share the same evaluation process under commonly-used metrics, the terms are sometimes used interchangeably in referenced papers.
Speaker recognition is also closely related to speaker diarization, where an input audio stream is partitioned into homogeneous segments according to the speaker identity \cite{garcia2017speaker}.

\textbf{\subsubsection{Voice Datasets}}
Some of the popular datasets on voice/speaker recognition are:

\textbf{NIST SRE:}
Starting in 1996, the National Institute of Standards and Technology (NIST) has organized a series of evaluations for speaker 
recognition research \cite{martin2001nist}. The \textbf{Speaker Recognition Evaluation (SRE) datasets} compiled by NIST have 
thus become the most widely used datasets for evaluation of speaker recognition systems. These datasets are collected in an
evolving fashion, and each evaluation plan has a slightly different focus. These evaluation datasets differ in audio lengths
\cite{nist2010sre}, recording devices (telephone, handsets, and video) \cite{nist2018sre}, data origination (in North America 
or outside) \cite{nist2016sre}, and match/mismatch scenarios. In recent years, \textbf{SRE 2016} and 
\textbf{SRE 2018} are the most popular datasets in this area.

\textbf{SITW:}
The \textbf{Speakers in the Wild (SITW)} dataset was acquired across unconstrained conditions \cite{mclaren2016speakers}. 
Unlike the SRE datasets, this data was not collected under controlled conditions and thus contains real noise and reverberation. 
The database consists of recordings of 299 speakers, with an average of eight different sessions per person.

\textbf{VoxCeleb:}
The \textbf{VoxCeleb dataset} \cite{nagrani2017voxceleb} and \textbf{VoxCeleb2 dataset} \cite{chung2018voxceleb2} are public 
datasets compiled from interview videos uploaded to YouTube to emphasize the lack of large scale unconstrained data for 
speaker recognition. These data are collected using a fully automated pipeline. A two-stream synchronization CNN is used to estimate the correlation between the audio track and the mouth motion of the video, and then
CNN-based facial recognition techniques are used to identify speakers for speech annotation. VoxCeleb1 contains over 100,000 
utterances for 1,251 celebrities, and VoxCeleb2 contains over a million utterances for 6,112 identities.

Apart from datasets designed purely for speaker recognition tasks, many datasets collected for automatic speech recognition 
can also be used for training or evaluation of speaker recognition systems. For example, the \textbf{Switchboard dataset} \cite{godfrey1997switchboard} and the \textbf{Fisher Corpus} \cite{cieri2004fisher}, which were originally collected for 
speech recognition tasks, are also used for model training in NIST Speaker Recognition Evaluations. On the other 
hand, researchers may utilize existing speech recognition datasets to prepare their own speaker recognition evaluation
dataset to prove the effectiveness of their research. For example, \textbf{Librispeech dataset} \cite{panayotov2015librispeech} 
and the \textbf{TIMIT dataset} \cite{zue1990speech} are pre-processed by the author in \cite{ravanelli2018learning} to serve as evaluation set for speaker recognition task.

\iffalse
Some of the popular datasets on voice recognition are:
\textbf{VoxCeleb:}
VoxCeleb is a large-scale speaker identification dataset. It contains around 100,000 phrases by 1,251 celebrities, extracted from YouTube videos, spanning a diverse range of accents, professions and age. “It’s an intriguing use case for isolating and identifying which superstar the voice belongs to,” according to VoxCeleb \cite{voxceleb}.

\textbf{TIMIT Dataset:}
The TIMIT corpus of read speech is designed to provide speech data for acoustic-phonetic studies and for the development and evaluation of automatic speech recognition systems. 
TIMIT contains broadband recordings of 630 speakers of eight major dialects of American English, each reading ten phonetically rich sentences. The TIMIT corpus includes time-aligned orthographic, phonetic and word transcriptions as well as a 16-bit, 16kHz speech waveform file for each utterance \cite{timit}.

\textbf{NIST SRE 2012:}
NIST Speaker Recognition Evaluation (SRE) consists of nine distinct tests. Each test uses one of three training conditions (core, telephone, microphone) and one of five test conditions (core, extended, summed,
known, and unknown). The benchmark uses 1,918 target speakers from speech corpora used
in previous SREs \cite{sre2012}.
\fi

\textbf{\subsubsection{Deep Learning Works on Voice Recognition}}
Before the era of deep learning, most state-of-the-art speaker recognition systems are built with the i-vectors approach \cite{dehak2010front},
which uses factor analysis to define a low-dimensional space that models both speaker and channel variabilities. 
In recent years, it has become more and more popular to explore deep learning approaches for speaker recognition. One of the 
first approach among these efforts is to incorporate DNN-based acoustic models into the i-vector framework \cite{lei2014novel}. 
This method uses an DNN acoustic model trained for Automatic Speech Recognition (ASR) to gather speaker statistics for i-vector model training. It has been 
shown that this improvement leads to a 30\% relative reduction in equal error rate.

Around the same time, d-vector was proposed in \cite{variani2014deep} to tackle text-dependent speaker recognition using neural network. 
In this approach, a DNN is trained to classify speakers at the frame-level. During enrollment and testing, the trained DNN is used to extract 
speaker specific features from the last hidden layer. “d-vectors” are then computed by averaging these features and used as speaker embeddings
for recognition. This method shows 14\% and 25\% relative improvement over an i-vector system under clean and noisy conditions, respectively.

In \cite{snyder2018x}, a time-delay neural network is trained to extract segment level “x-vectors” for text-independent speech recognition. 
This network takes in features of speech segments and passes them through a few non-linear layers followed by a
pooling layer to classify speakers at segment-level. X-vectors are then extracted from the pooling layer for enrollment and testing. 
It is shown that an x-vector system can achieve a better speaker recognition performance compared to the traditional i-vector approach, 
with the help of data augmentation.

End-to-end approaches based on neural networks are also explored in various papers. In \cite{heigold2016end} and \cite{zhang2016end}, neural  networks are designed to take in pairs of speech segments, and are trained to classify match/mismatch targets. A specially designed triplet  loss function is proposed in \cite{zhang2017end} to substitute a binary classification loss function. Generalized end-to-end (GE2E) loss, which is
similar to triplet loss, is proposed in \cite{wan2018generalized} for text-dependent speaker recognition on an in-house dataset.

In \cite{le2018robust}, a complementary optimizing goal called intra-class loss is proposed to improve deep speaker embeddings learned with 
triplet loss. It is shown in the paper that models trained using intra-class loss can yield a significant relative reduction of 30\% in equal error rate (EER) compared to the original triplet loss. The effectiveness is evaluated on both VoxCeleb and VoxForge datasets.

In \cite{ravanelli2018learning}, the authors proposed a method for learning speaker embeddings from raw waveform by maximizing the mutual information. 
This approach uses an encoder-discriminator architecture similar to that of Generative Adversarial Networks (GANs) to optimize mutual information
implicitly. The authors show that this approach effectively learns useful speaker representations, leading to a superior performance on the VoxCeleb 
corpus when compared with i-vector baseline and CNN-based triples loss systems.

In \cite{bhattacharya2019deep}, the authors combine a deep convolutional feature extractor, self-attentive pooling and large-margin loss functions
into their end-to-end deep speaker recognizers. The individual and ensemble models from this approach achieved state-of-the-art performance 
on VoxCeleb with a relative improvement of 70\% and 82\%, respectively, over the best reported results. The authors also proposed to use a neural 
network to subsitute PLDA classifier, which enables them to get the state-of-the-art results on NIST-SRE 2016 dataset.

\textbf{\subsection{Signature Recognition}}
Signature is considered a behavioral biometric. It is widely used in traditional and digital formats to verify the user's identity for the purposes of security, transactions, agreements, etc. Therefore, being able to distinguish an authentic signature from a forged one is of utmost importance.
Signature forgery can be performed as either a random forgery, where no attempt is made to make an authentic signature (e.g., merely writing the name \cite{radhika2011approach}), or a skilled forgery, where the signature is made to look like the original and is performed with the genuine signature in mind \cite{soleimani2016deep}.

In order to distinguish an authentic signature from a forged one, one may either store merely signature samples to compare against (offline verification), or also the features of the written signature such as the thickness of a stroke and the speed of the pen during the signing \cite{impedovo2008automatic}.
For verification, there are writer-dependent (WD) and  writer-independent (WI) methods. In WD methods, a classifier is trained for each signature owner, whereas, in WI methods, one is trained for all owners \cite{srihari2004learning}.

\textbf{\subsubsection{Signature Datasets}}
Some of the popular signature verification datasets include:

\textbf{ICDAR 2009 SVC:}
ICDAR 2009 Signature Verification Competition contains simultaneously acquired online and offline signature samples \cite{icdar_svc}. 
The online dataset is called "NFI-online" and was processed and segmented by Louis Vuurpijl. 
The offline dataset is called "NFI-offline" and was scanned by Vivian Blankers from the NFI. 
%The offline signatures were automatically segmented and visually inspected by Louis Vuurpijl. Processing of the segmented signatures was performed by Katrin Franke. 
The collection contains: authentic signatures from 100 writers, and forged signatures from 33 writers.
The NLDCC-online signature collection contains in total 1953 online and 1953 offline signature files.

\textbf{SVC 2004:}
Signature Verification Competition 2004 consists of two datasets for two verification tasks: one for pen-based input devices like PDAs and another one for digitizing
tablets \cite{SVC2004}. 
Each dataset consists of 100 sets of signatures with each set containing 20 genuine signatures and 20 skilled forgeries.

\textbf{Offline GPDS-960 Corpus:}
This offline signature dataset \cite{vargas2007off} includes signatures from 960 subjects. There are 24 authentic signatures for each person, and 30 forgeries performed by other people not in the original 960 (1920 forgers in total). Some works have used a subset of this public dataset, usually the images for the first 160 or 300 subjects, dubbing them GPDS-160 and GPDS-300 respectively.

\textbf{\subsubsection{Deep Learning Works on Signature Recognition}}
Before the rise of deep learning to its current popularity, there were a few works seeking to use it. For example, Ribeiro et al \cite{ribeiro2011deep} proposed a deep learning-based method to both identify a signature's owner and distinguish an authentic signature from a fake, making use of the Restricted Boltzmann Machine (RBM) \cite{ackley1985learning}.
With more powerful computer and massively parallel architectures making deep learning mainstream, the number of deep learning-based works increased dramatically, including those involving signature recognition.
Rantzsch et al \cite{rantzsch2016signature} proposed an embedding-based WI offline signature verification, in which the input signatures are embedded in a high-dimensional space using a specific training pattern, and the Euclidean distance between the input and the embedded signatures will determine the outcome.
Soleimani et al \cite{soleimani2016deep} proposed Deep Multitask Metric Learning (DMML), a deep neural network used for offline signature verification, mixing WD methods, WI methods, and transfer learning. Zhang et al \cite{zhang2016multi} proposed a hybrid WD-WI classifier in conjuction with a DC-GAN network in order to learn to extract the signature features in an unsupervised manner. 
\iffalse
The verification system is shown is Figure \ref{fig:zhang2016multi}.

\begin{figure}[h]
\begin{center}
   \includegraphics[page=16,width=0.7\linewidth]{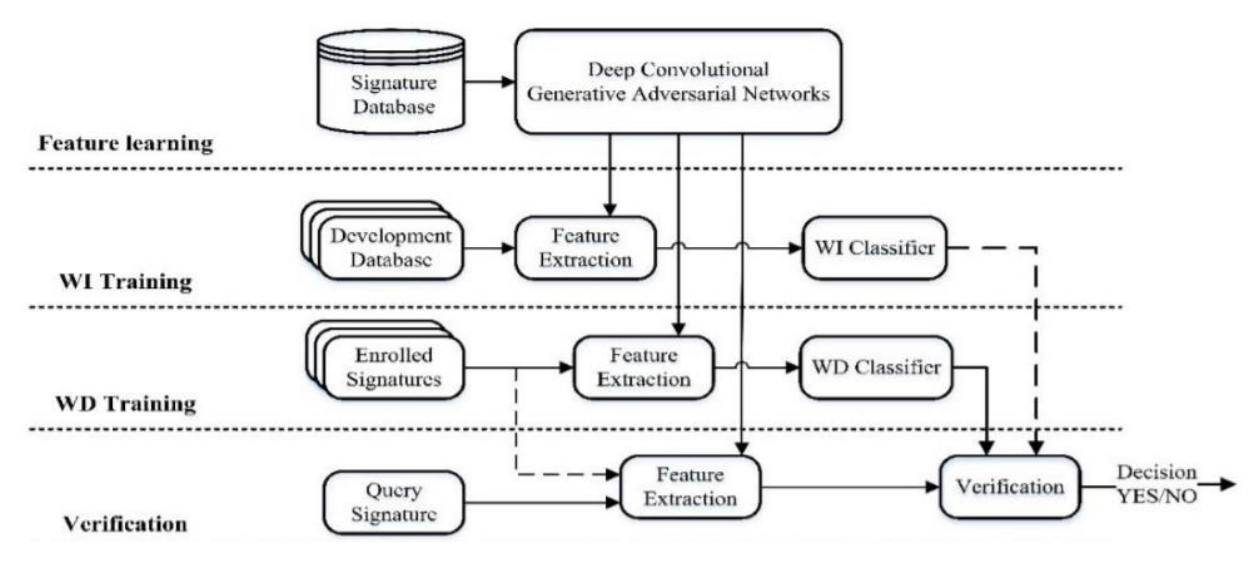}
\end{center}
   \caption{Multi-phase offline signature verification, courtesy of \cite{zhang2016multi}.}
\label{fig:zhang2016multi}
\end{figure}
\fi
With signature being a behavioral biometric, it is imperative to learn the best features to distinguish an authentic signature from a forged one.
Hafemann et al \cite{hafemann2017learning} proposed a WI CNN-based system to learn features of forgeries from multiple datasets, which greatly reduced the error equal rate compared to that of the state-of-the-art. 
Wang et al \cite{wang2019signature} proposed signature identification using a special GAN network (SIGAN) in which the loss value from the discriminator network is utilized as the threshold for the identification process.
Tolosana et al \cite{tolosana2018exploring} proposed an online writer-independent signature verification method using Siamese recurrent neural networks (RNNs), including long short term memory (LSTM) and gated recurrent units (GRUs).

\textbf{\subsection{Gait Recognition}}
Gait recognition is a popular pattern recognition problem and attracts a lot of researchers from different communities such as computer vision, machine learning, biomedical, forensic studying and robotics. 
This problem has also great potential in industries such as visual surveillance, since gait can be observed from a distance without the need for the subject's cooperation.
Similar to other behavioral biometrics, it is difficult, however possible, to try to imitate someone else's gait \cite{zhang2016siamese}. It is also possible for the gait to change due to factors such as the carried load, injuries, clothing, walking speed, viewing angle, and weather conditions, \cite{alotaibi2017improved}, \cite{wolf2016multi}. It is also a challenge to recognize a person among a group of walking people \cite{chen2017multi}.
Gait recognition can be model-based, in which the the structure of the subject's body is extracted (meaning more compute demand), or appearance-based, in which features are extracted from the person's movement in the images \cite{wolf2016multi}, \cite{zhang2016siamese}. 

\textbf{\subsubsection{Gait Datasets}}
Some of the widely used gait recognition datasets include:

\textbf{CASIA Gait Database:}
This CASIA Gait Recognition Dataset contains 4 subsets: Dataset A (standard dataset) \cite{casia_gaitA}, Dataset B (multi-view gait dataset), Dataset C (infrared gait dataset), and Dataset D (gait and its corresponding footprint dataset) \cite{CASIA_gait}.
Here we give details of CASIA B dataset, which is very popular. 
Dataset B is a large multi-view gait database, which is created in 2005. There are 124 subjects, and the gait data was captured from 11 views. 
Three variations, namely view angle, clothing and carrying condition changes, are separately considered. Besides the video files, they also provide human silhouettes extracted from video files.  The reader is referred to \cite{casia_gaitB} for more detailed information about Dataset B.

\textbf{Osaka Treadmill Dataset:}
This dataset has been collected in March 2007 at the Institute of Scientific and Industrial Research (ISIR), Osaka University (OU) \cite{osaka_url}. 
The dataset consists of 4,007 persons walking on a treadmill surrounded by the 25 cameras at 60 fps, 640 by 480 pixels.
The datasets are basically distributed in a form of silhouette sequences registered and size-normalized to 88x128 pixels size. 
They have four subsets of this dataset, dataset A: Speed variation, dataset B: Clothes variation, dataset C: view variations, and dataset D: Gait fluctuation. 
The dataset B is composed of gait silhouette sequences of 68 subjects from the side view with clothes variations of up to 32 combinations.
Detailed descriptions about all these datasets can be found in this technical note \cite{osaka_gait}.

\textbf{Osaka University Large Population (OULP) Dataset:}
This dataset \cite{iwama2012isir} includes images from 4,016 subjects from different ages (up to 94 years old) taken from two surrounding cameras and 4 observation angles. The images are normalized to 88x128 pixels.

\textbf{\subsubsection{Deep Learning Works on Gait Recognition}}
Research on gait recognition based on deep learning has only taken off in the past few years. In one of the older works, Wolf et al \cite{wolf2016multi} proposed a gait recognition system using 3D convolutional neural networks which learns the gait from multiple viewing angles. 
This model consists of multiple layers of 3D convolutions, max pooling and ReLUs, followed by fully-connected layers.
\iffalse
Its network architecture is shown in Figure \ref{fig:wolf2016multi}.
\begin{figure*}[h]
\begin{center}
   \includegraphics[page=17,width=0.8\linewidth]{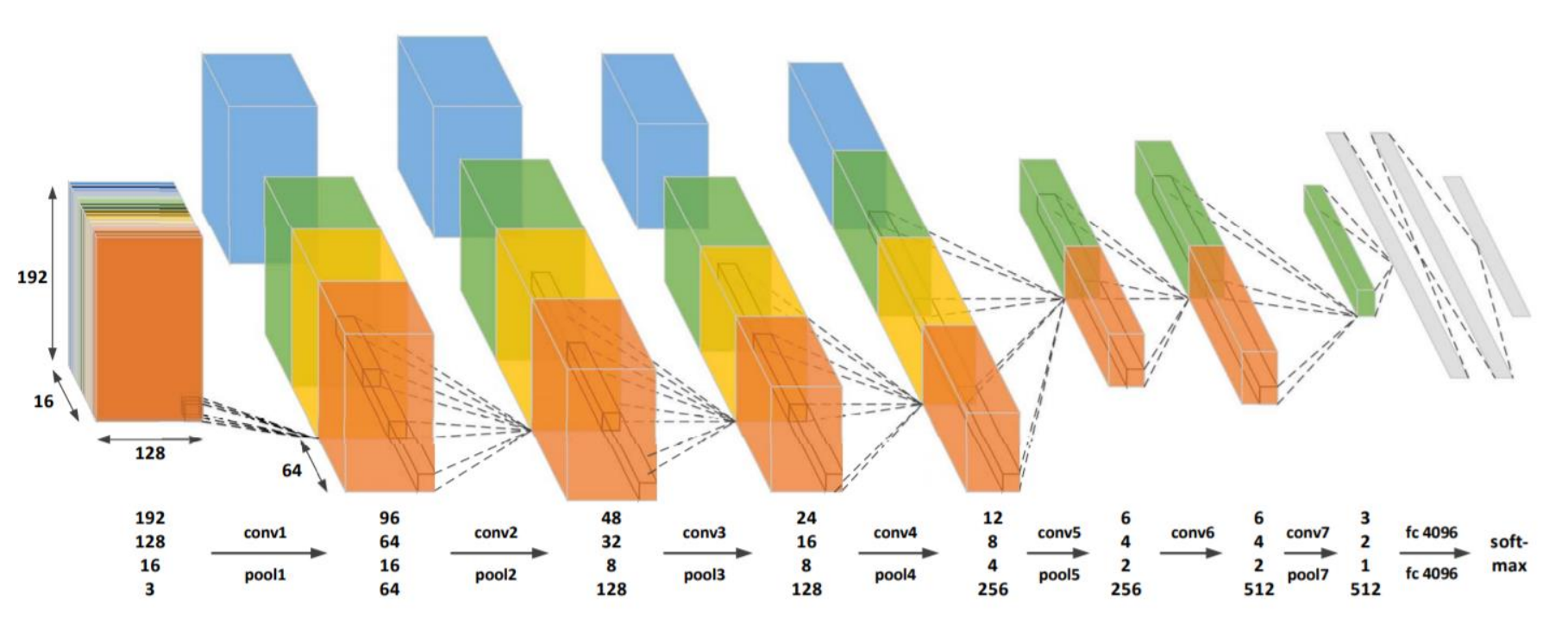}
\end{center}
   \caption{Network architecture of the multi-view gait recognition, courtesy of \cite{wolf2016multi}. It consists of multiple layers of 3D convolutions, max pooling and ReLUs, followed by fully-connected layers. Dimensions and number of layers are shown under each level.}
\label{fig:wolf2016multi}
\end{figure*}
\fi

\begin{figure*}[h]
\begin{center}
   \includegraphics[page=23,width=0.75\linewidth]{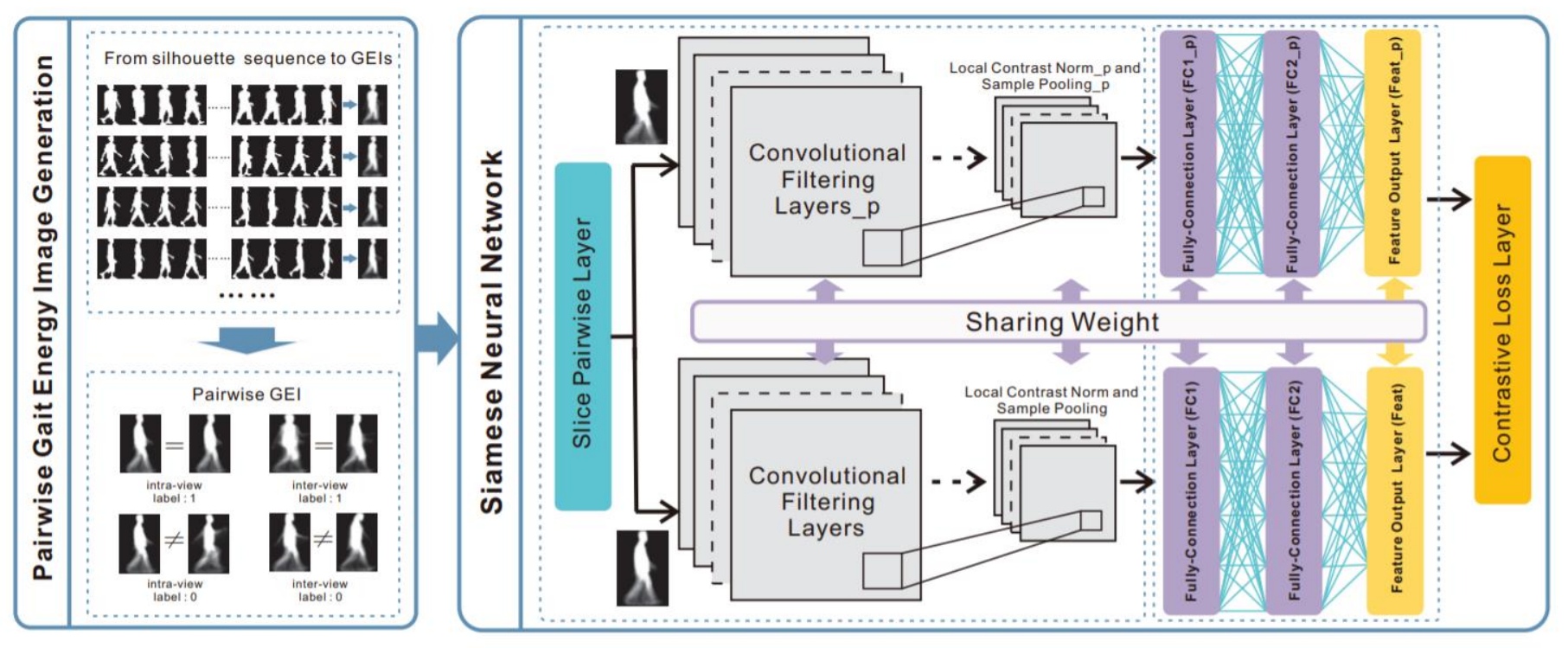}
\end{center}
   \caption{Siamese network for gait recognition, courtesy of \cite{zhang2016siamese}.}
\label{fig:zhang2016siamese}
\end{figure*}

Zhang et al \cite{zhang2016siamese} proposed a Siamese neural network for gait recognition, in which the sequences of images are converted into gait energy images (GEI) \cite{han2005individual}. Next, they are fed to the twin CNN networks and their contrastive losses are also calculated. This allows the system to minimize the loss for similar inputs and maximize it for different ones. 
The network for this work is shown in Figure \ref{fig:zhang2016siamese}.
Battistone et al \cite{battistone2019tglstm} proposed gait recognition through a time-based graph LSTM network, which uses alternating recursive LSTM layers and dense layers to extract skeletons from the person's images and learn their joint features.
Zou et al \cite{zou2018deep} proposed a hybrid CNN-RNN network which uses the data from smartphone sensors for gait recognition, particularly from the accelerometer and the gyroscope, and the subjects are not restricted in their walking in any way.

\vspace{0.8cm}
\section{Performance of Different Models on Different Datasets}
In this section, we are going to present the performance of different biometric recognition models developed over the past few years. 
We are going to present the results of each biometric recognition model separately, by providing the performance of several promising works on one or two widely used dataset of that biometric.
Before getting into the quantitative analysis, we are going to first briefly introduce some of the popular metrics that are used for evaluating biometric recognition models.

\textbf{\subsection{Popular Metrics For Evaluating Biometrics Recognition Systems}}
Various metrics are designed to evaluate the performance a biometric recognition systems. 
Here we provide an overview of some of the popular metrics for evaluation verification and identification algorithms.

\textbf{Biometric verification} is relevant to the problem of re-identification, where we want to see if a given data matches a  registered sample.
In many cases the performance is measured in terms of verification accuracy, specially when a test dataset is provided.
Equal error rate (EER) is another popular metric, which is the rate of error decided by a threshold that yields equal false negative rate and false positive rate. 
Receiver operating characteristic (ROC) is also another classical metric used for verification performance.
ROC essentially measures the true positive rate (TPR), which is the fraction of genuine comparisons that correctly exceeds the threshold, and the false positive rate (FPR), which is the fraction of impostor comparisons that incorrectly exceeds the threshold, at different thresholds.
ACC (classification accuracy) is another metric used by LFW, which is simply the percentage of correct classifications.
Many works also use TPR for a certain FPR.
For example  IJB-A focuses TPR@FAR=$10^{-3}$, while Megaface uses TPR@FPR= $10^{-6}$.

\textbf{Closed-set identification} can be measured in terms of closed-set identification accuracy, as well as rank-N detection and identification rate.
Rank-N measures the percentage of probe searches return the samples from probe’s gallery  within the top N rank-ordered results (e.g. IJB-A/B/C focuses on the rank-1 and rank-5 recognition rates). 
The cumulative match characteristic (CMC) is another popular metric, which measures the percentage of probes identified within a given rank. 
Confusion matrix is also a popular metric for smaller datasets.

\textbf{Open-set identification} deals with the cases where the recognition system should reject unknown/unseen subjects (probes which are not present in gallery) at the test time. 
At present, there are very few databases covering the task of open-set biometric recognition.
Open-set identification accuracy is a popular metrics for this task.
Some benchmarks also suggested to use the decision error trade-off (DET) curve to characterize the
FNIR (false-negative identification rate) as a function of FPIR (false-positive identification rate).

\noindent\textbf{Performance of Models for Face Recognition:}
For face recognition, various metrics are used for verification and identification.
For face verification, EER is one of the most popular metrics. 
For identification, various metrics are used such as close-set identification accuracy, open-set identification accuracy.
For open-set performance, many works used detection and identification accuracy at a certain false-alarm rate (mostly 1\%).

Due to the popularity of face recognition, there are a large number of algorithms and datasets available. 
Here, we are going to provide the performance of some of the most promising deep learning-based face recognition models, and their comparison with some of the promising classical face recognition models on three popular datasets. 

As mentioned earlier, LFW is one of the most widely used for face recognition. 
%Model performance can be measure under two different training settings: image-restricted and unrestricted. Under the image-restricted setting, only binary "matched" or "mismatched" labels are given for pairs of images.  Under the unrestricted setting, the identity information of the person appearing in each image is also available, allowing one to potentially form additional image pairs. 
%Some models make use of external images while to improve the performance.
The performance of some of the most prominent deep learning-based face verification models on this dataset is provided in Table~\ref{tab:face_lfw}.
We have also included the results of two very well-known classical face verification works. 
As we can see, models based on deep learning algorithms achieve superior performance over classical techniques with a large margin.
In fact, many deep learning approaches have surpassed human performance and are already close to 100\% (For verification task, not identification).

% results from below and 3 other papers
% https://arxiv.org/pdf/1804.06655.pdf
\begin{table*}[ht]
\centering
  \caption{Accuracy of different face recognition models for face verification on LFW dataset.}
  \centering
\begin{tabular}{|c|c|c|c|}
\hline
Method & Architecture  & Used Dataset & Accuracy on LFW \\ \hline
\textbf{Joint Bayesian \cite{jointbayes}}  & Classical   & -   & 92.4 \\ \hline
\textbf{Tom-vs-Pete \cite{Tom-vs-Pete}}  & Classical   & -  & 93.3  \\ \hline
\textbf{DeepFace \cite{DeepFace}}  & AlexNet   & Facebook (4.4M,4K)   & 97.35  \\ \hline
\textbf{DeepID2 \cite{DeepID2}} & AlexNet   & CelebFaces+ (0.2M,10K)   & 99.15  \\ \hline
\textbf{VGGface \cite{VGGface_model} } & VGGNet-16 & VGGface (2.6M,2.6K)   & 98.95  \\ 
\hline
\textbf{DeepID3 \cite{DeepID3}} & VGGNet-10 & CelebFaces+ (0.2M,10K)   & 99.53  \\ \hline
\textbf{FaceNet \cite{FaceNet}} & GoogleNet-24   & Google (500M,10M)   & 99.63  \\ \hline
\textbf{Range Loss \cite{RangeLoss}} & VGGNet-16   & MS-Celeb-1M, CASIA-WebFace  & 99.52  \\ \hline
\textbf{L2-softmax \cite{L2softmax}} & ResNet-101   & MS-Celeb-1M (3.7M,58K)   & \textbf{99.87}  \\ \hline
%\textbf{CoCo loss \cite{cocoloss}} & -   & MS-Celeb-1M (3M,80K)   & 99.86  \\ \hline
\textbf{Marginal Loss \cite{marginalloss}} & ResNet-27   & MS-Celeb-1M (4M,80K)   & 99.48  \\ \hline
\textbf{SphereFace \cite{Sphereface}} & ResNet-64 & CASIA-WebFace (0.49M,10K)   & 99.42  \\ \hline
\textbf{AMS loss \cite{amsloss}} & ResNet-20 & CASIA-WebFace (0.49M,10K)   & 99.12  \\ \hline
\textbf{Cos Face \cite{CosFace}} & ResNet-64 &  CASIA-WebFace (0.49M,10K)  & 99.33  \\ \hline
\textbf{Ring loss \cite{RingLoss}} & ResNet-64 & CelebFaces+ (0.2M,10K)   & 99.50  \\ \hline
\textbf{Arcface \cite{Arcface}} & ResNet-100 & MS-Celeb-1M (3.8M,85K)   & 99.45  \\ \hline
\textbf{AdaCos \cite{adacos}} & ResNet-50 & WebFace  & 99.71  \\ \hline
\textbf{P2SGrad \cite{p2sgrad}} & ResNet-50 & CASIAWebFace   & \underline{99.82}  \\ \hline
\end{tabular}
\label{tab:face_lfw}
\end{table*}

%\usepackage[utf8]{inputenc}
%\DeclareUnicodeCharacter{2212}{-}

%% verification results from this paper
% http://openaccess.thecvf.com/content_CVPR_2019/papers/Liu_AdaptiveFace_Adaptive_Margin_and_Sampling_for_Face_Recognition_CVPR_2019_paper.pdf
As mentioned earlier, closed-set identification is another popular face recognition task.
Table~\ref{tab:face_megaface}, provides the summary of the performance of some of the recent state-of-the-art deep learning-based works on the MegaFace challenge 1 (for both identification and verification tasks).
MegaFace challenge evaluates rank1 recognition rate as a function of an increasing number of gallery distractors (going from 10 to 1 million) for identification accuracy.
For verification, they report TPR at FAR= $10^{-6}$.
Some of these reported accuracies are taken from \cite{adapface}, where they  implemented the Softmax, A-Softmax, CosFace, ArcFace and the AdaptiveFace models with the same 50-layer CNN, for fair comparison.
As we can see the deep learning-based models in recent years achieve very high Rank-1 identification accuracy even in the case where 1 million distractors are included in the gallery.

\begin{table*}[ht]
\centering
  \caption{Face identification and verification evaluation on MegaFace Challenge 1.}
  \centering
\begin{tabular}{|c|c|c|c|}
\hline
\shortstack{ Method   \\ ~} &
\shortstack{ Protocol \\ ~} &
\shortstack{ Rank1 Identification   \\ Accuracy} &
\shortstack{ (TPR@$10^{-6}$FPR) \\ Verification Accuracy} \\
\hline
\textbf{Beijing FaceAll Norm 1600, from \cite{adapface}}  & Large   & 64.8  & 67.11  \\ \hline
\textbf{Softmax  \cite{adapface}} & Large & 71.36 & 73.04  \\ \hline
\textbf{Google - FaceNet v8 \cite{FaceNet}} & Large   &  70.49  & 86.47  \\ \hline
\textbf{YouTu Lab, from \cite{adapface}} & Large & 83.29   & 91.34  \\ \hline
\textbf{DeepSense V2, from \cite{adapface}} & Large & 81.29   & 95.99  \\ \hline
\textbf{Cos Face (Single-patch) \cite{CosFace}} & Large & 82.72   & \underline{96.65}  \\ \hline
\textbf{Cos Face (3-patch ensemble) \cite{CosFace}} & Large & 84.26    &  \textbf{97.96}  \\ \hline
\textbf{SphereFace \cite{Sphereface}} & Large & 92.241 & 93.423  \\ \hline
\textbf{Arcface \cite{Arcface}} & Large &   \underline{94.637} &  94.850  \\ \hline
\textbf{AdaptiveFace \cite{adapface}} & Large & \textbf{95.023} & 95.608  \\ \hline
\end{tabular}
\label{tab:face_megaface}
\end{table*}

Deep learning-based models have achieved great performance on other facial analysis tasks too, such as facial landmark detection, facial expression recognition, face tracking, age prediction from face, face aging, part of face tracking, and many more. 
As this paper is mostly focused on biometric recognition, we skip the details of models developed for those works here.

\iffalse
\begin{table}
\centering
  \caption{Performance of state-of-the-art methods on MegaFace Challenge 1.}
  \centering
\begin{tabular}{|c|c|c|}
\hline
& \multicolumn{2}{|c|}{MegaFace Challenge 1.  
FaceScrub} \\  
\hline
Method & Rank1@$10^6$ \ \ \ \ \ \ & TPR@$10^{-6}$FPR \\
\hline
Arcface \cite{Arcface} &  0.9836 & 0.9848 \\  
\hline
Cosface \cite{CosFace} &  0.9833 & 0.9841 \\  
\hline
Sphereface \cite{Sphereface} &  0.9743 & 0.9766 \\
\hline
Marginal Loss \cite{marginalloss} &  0.8028 & 0.9264 \\  
\hline
\end{tabular}
\label{tab:face_megaface}
\end{table}
\fi

\noindent\textbf{Performance of Models for Fingerprint Recognition:}
It is common for fingerprint recognition models to report their results using either the accuracy or equal error rate (EER).
Table \ref{tab:fingerprint} provides the accuracy of some of the recent fingerprint recognition works on PolyU, FVC, and CASIA databases. 
As we can see, deep learning-based models achieve very high accuracy rate on these benchmarks.

\begin{table}[ht]
  \centering
  \caption{Accuracy of several fingerprint recognition algorithms.}
\csvreader[tabular=|c|c|c|,
  no head,column count=3,
  table head=\hline,
  late after line=\\\hline]%
  {csv/fingerprint.csv}{}%
  {\csvlinetotablerow}%
  \label{tab:fingerprint}
\end{table}

\noindent\textbf{Performance of Models for Iris Recognition:}
Many of the recent iris recognition works have reported their accuracy rates on different iris databases, making it hard to compare all of them on a single benchmark.
The performance of deep learning-based iris recognition algorithms, and their comparison with some of the promising classical iris recognition models are provided in Table~\ref{tab:iris_table}.
As we can see models based on deep learning algorithms achieve superior performance over classical techniques.
Some of these numbers are taken from \cite{iris_survey} and \cite{iris_2019}.

% 1- https://www.researchgate.net/profile/J_Winston/publication/327349604_A_comprehensive_review_on_iris_image-based_biometric_system/links/5be987bda6fdcc3a8dd0da3b/A-comprehensive-review-on-iris-image-based-biometric-system.pdf
% 2- https://reader.elsevier.com/reader/sd/pii/S1319157818313302?token=D84E4B4EBD2BBC64A362A353F1DF12A2A2131309536710F177FEEE33440D5115F977F3CE45A68782279DE5C441251BA3
\begin{table*}[ht]
  \centering
    \caption{The performance of iris recognition models on some of the most popular datasets.}
\begin{tabular}{|c|c|c|c|}
\hline
Method    & Dataset & Model/Feature & Performance \\ \hline
\textbf{Elastic Graph Matching \cite{elastic}}  & IITD  & -   & Acc= 98\%  \\ 
\hline %\shortstack{CASIA, MMU, \\ UBIRIS}
\shortstack{\textbf{SIFT Based Model \cite{sift_iris}} \\ ~} &
\shortstack{ CASIA, MMU,  \\ UBIRIS} &
\shortstack{ SIFT features \\ ~} &
\shortstack{Acc= 99.05\% \\ EER=3.5\% } 
\\ \hline
\textbf{Deep CNN \cite{menon2018iris}} & IITD & - & Acc=99.8\%
\\ \hline
\textbf{Deep CNN \cite{menon2018iris}} & UBIRIS v2 & - & Acc=95.36\%
\\ \hline
%\textbf{Moment Features \cite{}} & CASIA, UPOL, UBIRIS  & -  & Acc= 97.5\%, EER=0.1\%    \\ \hline
\textbf{Deep Scattering \cite{iris_0}} & IITD   & ScatNet3+Texture features  & Acc= 99.2\%    \\ 
\hline
\textbf{Deep Features \cite{iris_1}} & IITD   & VGG-16  & Acc= 99.4\%    \\ 
\hline
\shortstack{\textbf{SCNN \cite{iris_scnn}} \\ ~} & \shortstack{CASIA-v4, FRGC \\ FOCS}   & \shortstack{Semantics-assisted \\ convolutional networks}  & \shortstack{R1-ACC= 98.4  \\ (CASIA-v4)}    \\ 
\hline
\end{tabular}

%\end{subtable}
%\caption{Cold start experiments.}
\label{tab:iris_table}
\end{table*}

\noindent\textbf{Performance of Models for Palmprint Recognition:}
It is common for palmprint recognition papers to compare their work against others using the accuracy rate or equal error rate (EER). Table \ref{tab:palmprint} displays the accuracy of some of the palmprint recognition works.
As we can see, deep learning-based models achieve very high accuracy rate on PolyU palmprint dataset.
\begin{table}[ht]
  \centering
  \caption{Accuracy of various palmprint recognition systems.}
\csvreader[tabular=|c|c|c|,
  no head,column count=3,
  table head=\hline,
  late after line=\\\hline]%
  {csv/palmprint.csv}{}%
  {\csvlinetotablerow}%
  \label{tab:palmprint}
\end{table}

\noindent\textbf{Performance of Models for Ear Recognition:}
The results of some of the recent ear recognition models are provided in Table \ref{tab:ear}. 
Besides recognition accuracy, some of the works have also reported their rank-5 accuracy, i.e. if one of the first 5 outputs of the algorithm is correct, the algorithm has succeeded.
Different deep learning-based models for ear recognition report their accuracy on different benchmarks.
Therefore, we list some of the promising works, along with the respective datasets that they are evaluated on, in  Table \ref{tab:ear}.

\begin{table}[ht]
  \centering
  \caption{Accuracy of select ear recognition algorithms.}
\csvreader[tabular=|c|c|c|,
  no head,column count=3,
  table head=\hline,
  late after line=\\\hline]%
  {csv/ear.csv}{}%
  {\csvlinetotablerow}%
  \label{tab:ear}
\end{table}

\noindent\textbf{Performance of Models for Voice Recognition:}
The most widely used metric for evaluation of speaker recognition systems is Equal Error Rate (EER).
Apart from EER, other metrics are also used for system evaluation. For example, detection error trade-off curve (DET curve) is used in SRE performance evaluations
to compare different systems. A DET curve is created by plotting the false negative rate versus false positive rate, with logarithmic 
scale on the x- and y-axes. (EER corresponds to the point on a DET curve where false negative rate and false positive rate are equal.)
Minimum detection cost is another metric that is frequently used in speaker recognition tasks \cite{van2007introduction}. This cost
is defined as a weighted average of two normalized error rates. Not all of these metrics are reported in every research papers, but
EER is the most important metric to compare different systems.

Table~\ref{tab:sre} records the performance of some of the best deep leaning based spearker recognition systems on VoxCeleb1
dataset. As is shown in the table, the progress made by researchers over the last two years are prominent. All these systems shown in 
Table~\ref{tab:sre} are single systems, which means the performance can be boosted further with system combination or ensembles.
\begin{table*}[thbp!]
    \centering
 \vspace{-0.8em}
  \caption{\label{tab:sre} {Accuracy of different speaker recognition systems on VoxCeleb1 dataset}}
    \begin{tabular}{ | c | c | c | c | }
    \hline
      Model  & Loss & Training set &  \ \ EER \ \ \\ 
    \hline
    \textbf{i-vector + PLDA \cite{shon2018frame}} & - &  VoxCeleb1 & 5.39 \\
    \hline
    \textbf{SincNet+LIM (raw audio) \cite{ravanelli2018learning}} & - & VoxCeleb1 & 5.80 \\
    \hline
    \textbf{x-vector* \cite{shon2018frame}} & Softmax & VoxCeleb1 &  6.00 \\
    \hline
    \textbf{ResNet-34 \cite{cai2018exploring}} &  A-Softmax + GNLL & VoxCeleb1 & 4.46  \\
    \hline
    \textbf{x-vector  \cite{okabe2018attentive}} & Softmax & VoxCeleb1 & 3.85 \\
    \hline
    \textbf{ResNet-20 \cite{hajibabaei2018unified}} & AM-Softmax & VoxCeleb1 & 4.30 \\
    \hline
    \textbf{ResNet-50 \cite{chung2018voxceleb2}} & Softmax + Contrastive & VoxCeleb2 & 3.95 \\
    \hline
    \textbf{Thin ResNet-34 \cite{xie2019utterance}} & Softmax & VoxCeleb2 & \underline{3.22} \\
    \hline
    \textbf{ResNet-28 \cite{bhattacharya2019deep}} & AAM & VoxCeleb1 & \textbf{0.95} \\
    \hline
    \end{tabular}
\end{table*}

For SRE datasets, due to the large number of its series and complexity of different evaluation conditions, it is hard to compile all results
into one table. 
Also different papers may present results on different sets or conditions, making it hard to compare the performance across 
different approaches. 

The deep learning-based approaches discussed above have also been applied to other related areas, e.g. speech diarization, replay attack detection 
and language identification. Since this paper focuses on biometric recognition, we skip the details for these tasks.

\iffalse
\begin{table}[thbp!]
    \centering
 \vspace{-0.8em}
  \caption{\label{tab:sre} {\it Accuracy of different speaker recognition systems on SRE 12 Condition 2} }
    \begin{tabular}{ | c | c | c | }
    \hline
      Model  & Loss &  EER \\ 
    \hline
    i-vector + PLDA \cite{okabe2018attentive} & - & 1.50 \\
    \hline
    x-vector (attentive statistics pooling) \cite{okabe2018attentive} & Softmax & 1.47 \\
    \hline
    \end{tabular}
\end{table}
\fi

\noindent\textbf{Performance of Models for Signature Recognition:}
Most signature recognition works use EER as the performance metric, but sometimes, they also report accuracy. 
Table \ref{tab:signature_eer} summarizes the EER of several signature verification methods on GPDS dataset, where there are 12 authentic signature samples used for each person (except in \cite{soleimani2016deep} where it is 10 samples). 
In addition, Table \ref{tab:signature} provides the reported accuracy of a few other works on other datasets.

\begin{table}[ht]
  \centering
  \caption{Reported EER of selected signature recognition models on GPDS dataset (using 10-12 genuine samples).}
\csvreader[tabular=|c|c|c|,
  no head,column count=3,
  table head=\hline,
  late after line=\\\hline]%
  {csv/signature_eer.csv}{}%
  {\csvlinetotablerow}%
  \label{tab:signature_eer}
\end{table}

\begin{table}[ht]
  \centering
  \caption{Accuracy reported by some signature recognition models.}
\csvreader[tabular=|c|c|c|,
  no head,column count=3,
  table head=\hline,
  late after line=\\\hline]%
  {csv/signature.csv}{}%
  {\csvlinetotablerow}%
  \label{tab:signature}
\end{table}

\noindent\textbf{Performance of Models for Gait Recognition:}
Likely due to the different configurations of the existing gait datasets, it is difficult to compare the deep learning-based gait recognition works. The results are reported in the form of accuracies and EER across different gallery view angles and cross-view settings. 
For Gait recognition, it is common to compare rank-5 statistics as well as the normal rank-1 ones. We have gathered some of the averaged accuracy results reported in \cite{sundararajan2018deep} in Table \ref{tab:gait}. Note that results using CASIA-B are collected in various scene and viewing conditions, while, using OU-ISIR, they are for cross-view conditions.
\begin{table*}[ht]
  \centering
  \caption{Accuracy of select gait recognition models.}
\csvreader[tabular=|c|c|c|c|,
  no head,column count=4,
  table head=\hline,
  late after line=\\\hline]%
  {csv/gait.csv}{}%
  {\csvlinetotablerow}%
  \label{tab:gait}
\end{table*}

\vspace{0.8cm}
%%%%%%%%%%%%%%%%%%%%%%%%%%%%%%%%%%%%%%%%%%
\section{Challenges and Opportunities}
Biometric recognition systems have undergone great progress with the help of deep learning-based models, in the past few years, but there are still several challenges ahead which may be tackled in the few years and decades.

\textbf{\subsection{More Challenging Datasets}}
Although some of the current biometric recognition datasets (such as MegaFace, MS-Celeb-1M) contain a very large number of candidates, they are still far from representing all the real-world scenarios. 
Although state-of-the-art algorithms can achieve accuracy rates of over 99.9\% on LFW and Megaface databases, fundamental challenges such as matching faces/biometrics across ages, different poses, partial-data, different sensor types still remain challenging.
Also the number of subjects/people in real-world scenarios should be in the order of tens of millions.
Therefore biometrics dataset which contain a much larger number of classes (10M-100M), as well as a lot more intra-class variations, would be another big step towards supporting all real-world conditions. 

\textbf{\subsection{Interpretable Deep Models}}
It is true that deep learning-based models achieved an astonishing performance on many of the challenging benchmarks, but there are still several open questions about these models. For example, what exactly are deep learning models learning?
Why are these models easily fooled by adversarial examples (while human can detect many of those examples easily)?
What is a minimal neural architecture which can achieve a certain recognition accuracy on a given dataset?

\textbf{\subsection{Few Shot Learning, and Self-Supervised Learning}}
Many of the successful models developed for biometric recognition are trained on large datasets with enough samples for each class. 
One of the interesting future trend will be to develop recognition models which can learn a powerful models from very few shots (zero/one shot in extreme case). 
This would enable training discriminative models without the need to provide several samples for each person/identity.
Self-supervised learning \cite{selfsuper} is also another recent popular topic in deep learning, which has not been explored enough for biometrics recognition. 
One way to use it would be to learn discriminative biometric feature from local patches of an image, and then aggregating those features and used for classification.

\textbf{\subsection{Biometric Fusion}}
Single biometric recognition  by itself is far from
sufficient to solve all biometric/forensic tasks. 
For example distinguishing identical twins may not be possible from face only, or matching an identity from face with disguise, or after surgery may not be that easy. 
Fusing the information from multiple biometrics can provide a more reliable solution/system in many of these cases (for example using voice+face or voice+gait can potentially solve the identical twin detection) \cite{ross2004multimodal, ross2003information}.
A good neural architecture which can jointly encode and aggregate different biometrics would be an interesting problem (information fusion can happen at the data level, feature level, score level, or decision level). 
Image set classification could also be useful in this direction \cite{imageset}.
There have been some works on biometric fusion, but most of them are far from the real-world scale of biometric recognition, and are mostly in their infancy.
For some of the challenges of multi-modal machine learning, we refer the reader to \cite{multimodal_challenge}.

\textbf{\subsection{Realtime Models for Various Applications}}
In many applications, accuracy is the most important factor; however, there are many applications in which it is also crucial  to have a near real-time biometric recognition model.
This could be very useful for on-device solutions, such as the one for cellphone and tablet authentication.
Some of the current deep models for biometrics recognition are far from this speed requirement, and developing near real-time models yet accurate models would be very valuable.

\textbf{\subsection{Memory Efficient Models}}
Many of the deep learning-based models require a significant amount of memory even during inference.
So far, most of the effort has focused on improving the accuracy of these models, but in order to fit these models in devices, the networks must be simplified.
This can be done either by using a simpler model, using model compression techniques, or training a complex model and then using knowledge distillation techniques to compress that into a smaller network mimicking the initial complex model.
Having a memory-efficient model opens up the door for these models to be used even on consumer devices.

\textbf{\subsection{Security and Privacy Issues}}
Security is of great importance in biometric recognition systems.
Presentation attack, template attack, and adversarial
attack threaten the security of deep biometric recognition systems, and challenge the current anti-spoofing methods.
Although some attempts have been done for adversarial example detection, there is still a long way to robust/reliable anti-spoofing capabilities.

With the leakage of biological/biometrics data nowadays, privacy concerns are rising. 
Some information about the user identity/age/gender can be decoded from the neural feature representation of their images. Research on visual cryptography, to protect users’ privacy on stored biometrics templates are essential for addressing public concern on privacy.

\vspace{0.8cm}
%%%%%%%%%%%%%%%%%%%%%%%%%%%%%%%%%%%%%%%%%%
\section{Conclusions}
In this work, we provided a summary of the recent deep learning-based models (till 2019) for biometric recognition.
As opposed to the other surveys, it provides an overview of most used biometrics.
Deep neural models have shown promising improvement over classical models for various biometrics.
Some biometrics have attracted a lot more attention (such as face) due to the wider industrial applications, and availability of large-scale datasets, but other biometrics seem to be following the same trend.
Although deep learning research in biometrics has achieved promising results, there is still a great room for improvement in different directions, such as creating larger and more challenging datasets, addressing model interpretation, fusing multiple biometrics, and addressing security and privacy issues.

\iffalse
\begin{table}
\caption{Please write your table caption here}
\label{tab:1}       % Give a unique label
\begin{tabular}{lll}
\hline\noalign{\smallskip}
first & second & third  \\
\noalign{\smallskip}\hline\noalign{\smallskip}
number & number & number \\
number & number & number \\
\noalign{\smallskip}\hline
\end{tabular}
\end{table}
\fi

\begin{acknowledgements}
We would like to  thank Prof. Rama Chellappa, and Dr. Nalini Ratha for reviewing this work, and providing very helpful comments and suggestions.
\end{acknowledgements}



\bibliographystyle{plain}

\iffalse

\fi

\end{document}